\documentclass[oneside,11pt]{article}

\usepackage{times,url,mathrsfs}
\usepackage{amsmath,wrapfig,color}
\usepackage{amsfonts}
\usepackage{graphicx}
\usepackage{subfigure}
\usepackage{bbm}
\usepackage{MnSymbol}
\usepackage{mathtools}
\newtheorem{theorem}{Theorem}

\begin{document}
\title{\huge Asymmetric LSH (ALSH) for Sublinear Time Maximum Inner Product Search (MIPS) \vspace{0.2in}}

\author{
         \bf{Anshumali Shrivastava}\\
         Department of Computer Science\\
         Computer and Information Science\\
         Cornell University\\
         {Ithaca, NY 14853, USA}\\
        \texttt{anshu@cs.cornell.edu}
        \and
 \bf{Ping Li} \\
         Department of Statistics and Biostatistics\\\hspace{0.1in}
         Department of Computer Science\\
       Rutgers University\\
          Piscataway, NJ 08854, USA\\
       \texttt{pingli@stat.rutgers.edu}
        }

\date{}

\maketitle

\begin{abstract}
We\footnote{Initially submitted in  Feb. 2014.} present the first provably sublinear time algorithm for approximate \emph{Maximum Inner Product Search} (MIPS). Our proposal is also the first hashing algorithm for searching with (un-normalized) inner product as the underlying similarity measure. Finding hashing schemes for MIPS was considered hard. We formally show that the existing Locality Sensitive Hashing (LSH) framework is insufficient for solving MIPS, and  then we  extend the existing LSH framework to allow asymmetric hashing schemes. Our proposal is based on an interesting mathematical phenomenon in which inner products, after independent asymmetric transformations, can be converted into the problem of approximate near neighbor search. This key observation makes efficient sublinear hashing scheme for MIPS possible. In the extended asymmetric LSH (ALSH) framework, we provide an explicit construction of provably fast hashing scheme for MIPS. The proposed construction and the extended LSH framework could be of independent theoretical interest. Our proposed algorithm is simple and easy to implement. We evaluate the method, for retrieving inner products, in the collaborative filtering task of item recommendations on Netflix and Movielens  datasets.
\end{abstract}

\newpage\clearpage

\section{Introduction and Motivation}\label{sec:intro}
The  focus of this paper is on the problem of \emph{Maximum Inner Product Search (MIPS)}. In this problem, we are given a giant data vector collection $\mathcal{S}$ of size $N$, where $\mathcal{S} \subset \mathbb{R}^D$,  and a given query point $q \in \mathbb{R}^D$. We are interested in searching for  $p \in \mathcal{S}$ which maximizes (or approximately maximizes) the {\bf inner product $q^Tp$}. Formally, we are interested in efficiently computing
\begin{equation}
p = \arg\max_{x \in \mathcal{S}}\hspace{0.1in} q^Tx
\end{equation}
The MIPS problem is related to the problem of \emph{ near neighbor search (NNS)}, which instead requires computing
\begin{equation}
p = \arg\min_{x \in \mathcal{S}}||q -x||_2^2 =  \arg\min_{x \in \mathcal{S}} (||x||_2^2 -  2 q^Tx)
\end{equation}
These two problems are equivalent if the norm of every element $x \in \mathcal{S}$ is constant. Note that the value of the norm $||q||_2$ has no effect as it is a constant throughout and does not change the identity of $\arg\max$ or $\arg\min$.  There are many scenarios in which MIPS arises naturally at places where the norms of the elements in $\mathcal{S}$ have very significant variations~\cite{Proc:Koenigstein_CIKM12} and can not be controlled. As a consequence, existing fast algorithms for the problem of approximate NNS can not be directly used for solving MIPS.\\

Here we list a number of practical scenarios where the MIPS  problem is solved as a subroutine: (i) recommender system,   (ii) large-scale object detection with DPM, (iii) structural SVM, and (iv) multi-class label prediction.

\subsection{Recommender Systems}

Recommender systems are often based on collaborative filtering which relies on  past behavior of users, e.g., past  purchases and ratings. Latent factor modelling based on matrix factorization~\cite{Article:Koren_2009} is a popular approach for solving collaborative filtering. In a typical matrix factorization model, a  user $i$ is associated with a latent user characteristic vector $u_i$, and similarly, an item $j$ is associated with a latent item characteristic vector $v_j$.  The rating $r_{i,j}$ of item $j$ by user $i$ is modeled as the \textbf{inner product} between the corresponding  characteristic vectors. A popular generalization of this framework, which combines neighborhood information with latent factor approach~\cite{Proc:Koren_KDD2008}, leads to the following model:
\begin{align}\label{eq:svd++}
r_{i,j} = \mu + b_i + b_j + u_i^Tv_j
\end{align}
where $\mu$ is the over all constant mean rating value, $b_i$ and $b_j$ are user and item biases, respectively.  Note that Eq. (\ref{eq:svd++}) can also be written as $ r_{i,j} = \mu + [u_i;b_i;1]^T[v_j;1;b_j],$ where the form $[x;y]$ is the concatenation of vectors $x$ and $y$. \\

 Recently, \cite{Proc:Cremonesi_RecSys} showed that a simple computation of $u_i$ and $v_j$ based on naive SVD of the sparse rating matrix outperforms existing models, including the neighborhood model,  in recommending top-ranked items. In this setting, given a user $i$ and the corresponding learned latent vector $u_i$ finding the right item $j$, to recommend to this user, involves computing
 \begin{align}\label{eqn_argmax_r}
 j =\arg \max_{j^\prime}\ \ r_{i,j^\prime} = \arg\max_{j^\prime}\ \  u_i^Tv_{j^\prime}
  \end{align}
 which is an instance of the standard MIPS problem.  It should be noted that we do not have control over the norm of the learned characteristic vector, i.e., $\|v_j\|_2$, which often has a wide range in practice~\cite{Proc:Koenigstein_CIKM12}.

  If  there are $N$ items to recommend, solving (\ref{eqn_argmax_r}) requires computing $N$ inner products.  Recommendation systems  are typically deployed in on-line application over web where the number $N$ is huge. A brute force linear scan over all items, for computing  $\arg\max$, would be prohibitively expensive. 

%{\bf Feature Extraction Pipelines:} The use of \emph{Deformable Parts Model (DPM)} representation for images are popular and state-of-art in object detection task~\cite{}.  \\
\subsection{Large-Scale Object Detection with DPM}

Deformable Part Model (DPM) based representation of images is the state-of-the-art in object detection tasks~\cite{Article:Felzenszwalb_PAMI2010}. In DPM model, first a set of part filters are learned from the train dataset. During detection, these learned filter activations over various patches of the test image are used to score the test image. The activation of a filter on an image patch is an inner product between them.  Typically, the number of possible filters are large (e.g., millions) and so scoring the test image is  costly. Very recently, it was shown that scoring based only on filters with high activations performs well in practice~\cite{Proc:Dean_CVPR2013}. Identifying filters, from a large collection of possible filters, having high activations on a given image patch requires computing top inner products. Consequently,  an efficient solution  to the MIPS problem will  benefit large scale object detections based on DPM.

\subsection{Structural SVM}

Structural SVM, with cutting plane training~\cite{Article:joachims_2009}, is one of the popular methods for learning over structured data. The most expensive step with cutting plane iteration is the call to the separation oracle which identifies the most violated constraint. In particular, given the current SVM estimate $w$, the separation oracle computes
\begin{equation}\label{eqn_structuralSVM}
\hat{y}_i = \arg\max_{\hat{y} \in \mathcal{Y}}\ \ \Delta(y_i,\hat{y}) + w^T\Psi(x_i,\hat{y})
\end{equation}
where $\Psi(x_i,\hat{y})$ is the joint feature representation of data with the possible label $\hat{y}$ and $\Delta(y_i,\hat{y})$ is the loss function. Clearly, this is again an instance of the MIPS problem. This step is  expensive in that the number of possible elements, i.e., the size of $ \mathcal{Y}$, is possibly exponential. Many heuristics were deployed to hopefully improve the  computation of  $\arg\max$ in (\ref{eqn_structuralSVM}), for instance caching~\cite{Article:joachims_2009}. An efficient MIPS routine can make structural SVM faster and more scalable.

\subsection{Multi-Class Label Prediction}

The models for multi-class SVM (or logistic regression)  learn a weight vector $w_i$ for each of the class label $i$. After the weights are learned, given a new test data vector $x_{test}$, predicting its class label   is basically an MIPS problem:
\begin{equation}
y_{test} = \arg\max_{i\in \mathcal{Label}}\ \ x_{test}^T\ w_i
\end{equation}
where $\mathcal{Label}$ is the set of possible class labels. Note that the norms of the weight vectors $\|w_i\|_2$ are not constant.%, as typical in MIPS.

The size, $|\mathcal{Label}|$, of the set of class labels differs in applications. Classifying with large number of possible class labels is  common in fine grained object classification, for instance, prediction task with 100,000  classes~\cite{Proc:Dean_CVPR2013} (i.e., $|\mathcal{Label}|=100,000$). Computing such  high-dimensional vector multiplications for predicting the class label of a single instance can be expensive in, for example, user-facing applications.

%{\bf Greedy Coordinate Descent}   Greedy coordinate descent method greedily picks the coordinate, to perform a gradient step, that attains the maximum of the gradient vector~\cite{}. In particular, given a loss function over n training data $\{x_i,y_i\}_{i=1}^n$ \begin{equation}\mathcal{L}(w) = \sum_{i =1}^{n} l(w^Tx_i,y_i),\end{equation} to perform the gradient step co-ordinate wise, the simplest deterministic choice is the coordinate that maximizes \\

\subsection{The Need for Hashing Inner Products}

Recall the MIPS problem is  to find $x\in\mathcal{S}$ which maximizes the inner product between $x$ and the given query $q$, i.e. $\max_{x \in \mathcal{S}}\ q^Tx$. A brute force scan of all elements of $\mathcal{S}$ can be prohibitively costly in applications which deal with massive data and care about the latency (e.g., search).

Owing to the significance of the problem, there was an attempt to efficiently solve MIPS by making use of tree data structure combined with branch and bound space partitioning technique~\cite{Proc:Ram_KDD12,Proc:Koenigstein_CIKM12} similar to k-d trees~\cite{Article:Friedman_74}. That  method  did not come with  provable runtime guarantees. In fact, it is also well-known that techniques based on space partitioning  (such as k-d trees) suffer from the curse of dimensionality. For example, it was shown in~\cite{Proc:Weber_VLDB98} (both empirically and theoretically) that all current  techniques (based on space partitioning) degrade to linear search, even for dimensions as small as 10 or 20.

 Locality Sensitive Hashing (LSH)~\cite{Proc:Indyk_STOC98}  based randomized techniques are common and successful in industrial practice for efficiently solving NNS ({\em near neighbor search}). Unlike space partitioning techniques, both the running time as well as the accuracy guarantee of LSH based NNS are in a way independent of the dimensionality of the data. This makes LSH suitable for large scale processing system dealing with ultra-high dimensional datasets which are  common these days.  Furthermore, LSH based schemes are massively parallelizable, which makes them ideal for modern ``Big'' datasets. The prime focus of this paper will be on  efficient hashing based algorithms for MIPS, which do not suffer from the curse of dimensionality.

\subsection{Our Contributions}

We develop {\em Asymmetric LSH (ALSH)}, an extended LSH scheme for efficiently solving the approximate MIPS problem.  Finding hashing based algorithms for MIPS was considered hard~\cite{Proc:Ram_KDD12,Proc:Koenigstein_CIKM12}. In this paper, we formally show that this is indeed the case with the current framework of LSH, and there can not exist any LSH for solving MIPS.  Despite this negative result, we show that it is possible to relax the current LSH framework to allow asymmetric hash functions which can efficiently solve MIPS. This generalization comes with no cost and the extended framework inherits all the theoretical  guarantees of LSH.

Our construction of asymmetric LSH is based on an interesting mathematical phenomenon that the original MIPS problem, after  asymmetric transformations, reduces to the problem of approximate near neighbor search. Based on this key observation, we show an explicit construction of asymmetric hash function, leading to the first provably sublinear query time algorithm for approximate similarity search with (un-normalized) inner product as the similarity. The construction of asymmetric hash function and the new LSH framework could be of independent theoretical interest.

Experimentally, we evaluate our algorithm for the task of recommending top-ranked  items, under the collaborative filtering framework, with the proposed asymmetric hashing scheme on Netflix and Movielens datasets. Our evaluations support the theoretical results and clearly show that the proposed asymmetric hash function is superior for retrieving inner products, compared to the well known hash function based on p-stable distribution for L2 norm~\cite{Proc:Datar_SCG04} (which is also part of standard LSH package~\cite{Report:E2LSH}). This is not surprising because L2 distances and inner products may have very different orderings.

\section{Background}\label{sec:allhashes}

%Early attempts of finding algorithms with sublinear runtime performance for exact near neighbor search, based on space partitioning, turned out to be a disappointment with the massive dimensionality of datasets \cite{WeberSB98}.

\subsection{ Locality Sensitive Hashing (LSH)}

Approximate versions of the near neighbor search problem~\cite{Proc:Indyk_STOC98} were proposed to break the linear query time bottleneck. The following formulation is commonly adopted.\\

\noindent\textbf{Definition:} ($c$-Approximate Near Neighbor or $c$-NN)\ {\em Given a set of points in a $D$-dimensional space $\mathbb{R}^D$, and parameters $S_0 > 0$, $\delta > 0$, construct a data structure which, given any query point $q$, does the following with probability $1- \delta$: if there exists an $S_0$-near neighbor of $q$ in $P$, it reports some $cS_0$-near neighbor of $q$ in $P$.}\\

The usual notion of $S_0$-near neighbor is in terms of distance. Since we  deal with similarities, we can equivalently define $S_0$-near neighbor of point $q$ as a point $p$ with $Sim(q,p) \ge S_0$, where $Sim$ is the similarity function of interest.

The  popular technique for $c$-NN  uses the underlying theory of \emph{Locality Sensitive Hashing} (LSH)~\cite{Proc:Indyk_STOC98}. LSH is a family of functions, with the property that more similar input objects in the domain of these functions have a higher probability of colliding in the range space than less similar ones. In formal terms, consider $\mathcal{S}$ a family of hash functions mapping $\mathbb{R}^D$ to some set $\mathcal{I}$.\\
%\begin{mydef}
%\label{def:LSH}

\noindent\textbf{Definition:} (Locality Sensitive Hashing (LSH))\ {\em A family $\mathcal{H}$ is called $(S_0,cS_0,p_1,p_2)$-sensitive if, for any two point $x,y \in \mathbb{R}^D$, $h$ chosen uniformly from $\mathcal{H}$ satisfies the following:
\begin{itemize}
\item if $Sim(x,y)\ge S_0$ then $Pr_\mathcal{H}(h(x) = h(y)) \ge p_1$
\item if $ Sim(x,y)\le cS_0$ then $Pr_\mathcal{H}(h(x) = h(y)) \le p_2$
\end{itemize}
For efficient approximate nearest neighbor search, $p_1 > p_2$ and $c < 1$ is needed.}

\subsection{Fast Similarity Search with LSH}
\label{sec:fastSimSearch}
In a typical task of similarity search, we are given a query  $q$, and our aim is to find $x \in \mathcal{S}$ with high value of $Sim(q,x)$. LSH provides a  clean mechanism of creating hash tables \cite{Report:E2LSH}. The idea is to concatenate $K$ independent hash functions to create a meta-hash function of the form
\begin{equation}\label{eq:bucket}
B_l(x)  = \left[h_{1}(x);h_{2}(x);...;h_{K}(x)\right]
\end{equation}
 where $ h_i, i = \{1,2,...,K \}$ are $K$ independent  hash functions sampled from the LSH family. The LSH algorithm  needs $L$ independent meta hash functions $B_l(x)$, $l=1, 2,..., L$.
\begin{itemize}
\item {\bf Pre-processing Step:} During preprocessing, we assign $x_i \in \mathcal{S}$ to the bucket $B_l(x_i)$ in the hash table $l$, for $l= 1, 2, ..., L$.
\item {\bf Querying Step:} Given a query $q$, we retrieve union of all elements from buckets $B_l(q)$, where the union is taken over all hash tables $l$, for $l= 1, 2, ..., L$.
\end{itemize}
The bucket $B_l(q)$ contains elements $x_i \in \mathcal{S}$ whose $K$ different hash values collide with that of the query. By the LSH property of the hash function, these elements have higher probability of being similar to the query $q$ compared to a random point. This probability value can be tuned by choosing appropriate value for parameters $K$ and $L$. Optimal choices lead to fast query time algorithm:\\

\noindent\textbf{Fact 1}:\ Given a family of $(S_0,cS_0,p_1,p_2)$ -sensitive hash functions, one can construct a data structure for $c$-NN with $O(n^\rho \log{n})$ query time and space $O(n^{1 + \rho})$, where $\rho = \frac{\log{p_1}}{\log{p_2}} <1$.\\

LSH trades off query time with extra (one time) preprocessing  cost and space.  Existence of an LSH family translates into provably sublinear query time algorithm for c-NN problems. It should be noted that the worst case query time for LSH is only dependent on $\rho$ and $n$. Thus, LSH based near neighbor search in a sense does not suffer from the curse of dimensionality. This makes LSH a widely popular technique in industrial practice~\cite{Proc:Henzinger_SIGIR06,Proc:Manku_WWW07,Proc:Das_WWW07}.

\subsection{LSH for L2 distance}

\label{sec:L2Hash}
\cite{Proc:Datar_SCG04} presented a novel LSH family for all $L_p$  ($p \in (0,2]$) distances. In particular, when $p =2$, this scheme provides an LSH family for $L_2$ distances.  Formally, given a fixed number $r$, we choose  a random vector $a$ with each component generated from i.i.d. normal, i.e., $a_i \sim N(0,1)$, and a scalar $b$ generated uniformly at random from $[0,r]$. The hash function  is defined as:
\begin{equation}\label{eq:L2Hash} h_{a,b}^{L2}(x) = \left\lfloor \frac{a^Tx + b}{r} \right\rfloor \end{equation}
where $\lfloor \rfloor$ is the floor operation. The collision probability under this scheme can be shown to be
\begin{align}\label{eq:L2collprob}
&Pr(h^{L2}_{a,b}(x) = h^{L2}_{a,b}(y)) = F_r(d),\\\label{eq:F}
&F_r(d) =  1 - 2\Phi(-r/d) - \frac{2}{\sqrt{2\pi}(r/d)}\left(1 - e^{-(r/d)^2/2}\right)
\end{align}
where $\Phi(x) = \int_{-\infty}^{x}\frac{1}{\sqrt{2\pi}}e^{-\frac{x^2}{2}}dx$ is the cumulative density function (cdf) of standard normal distribution and  $d = ||x -y||_2$ is the Euclidean distance between the vectors $x$ and $y$. This collision probability $F_r(d)$ is a monotonically decreasing function of the distance $d$ and hence  $h^{L2}_{a,b}$ is an LSH for $L2$ distances. This scheme is also the part of LSH package~\cite{Report:E2LSH}. Here $r$ is a parameter which can be tuned.

As argued previously, $||x -y||_2 = \sqrt{(||x||_2^2 +||y||_2^2 -2x^Ty)}$ is not monotonic in the inner product $x^Ty$ unless the given data has a constant norm.  Hence, $h^{L2}_{a,b}$ is not suitable for MIPS. In Section~\ref{sec:evaluations}, we will experimentally show  that our proposed method compares favorably to $h^{L2}_{a,b}$ hash function for MIPS.

Very recently~\cite{Report:RPCodeLSH2014} reported an improvement of this well-known hashing scheme when the data can be normalized (for example, when $x$ and $y$ both have unit $L_2$ norm). However, in our problem setting,  since the data can not be normalized,  we can not take advantage of the new results of~\cite{Report:RPCodeLSH2014}, at the moment.

\section{Hashing for MIPS}

\subsection{A Negative Result}

 We first show that, under the current LSH framework, it is impossible to obtain a locality sensitive hashing for MIPS. In~\cite{Proc:Ram_KDD12,Proc:Koenigstein_CIKM12}, the authors also argued that finding locality sensitive hashing for inner products could be hard, but to the best of our knowledge we have not seen a formal proof.
\begin{theorem}
There can not exist any LSH family for MIPS.\\

\noindent\textbf{Proof:}\ Suppose, there exists such hash function $h$. For un-normalized inner products the self similarity of a point $x$ with itself is  $Sim(x,x) =x^Tx = ||x||_2^2$ and there may exist another points $y$, such that $Sim(x,y) = y^Tx > ||x||_2^2 + M$, for any constant $M$.  Under any single randomized hash function $h$, the collision probability  of the event $\{h(x) = h(x)\}$ is always 1. So if $h$ is an LSH for inner product then the event $\{h(x) = h(y)\}$ should have higher probability compared to the event $\{h(x) = h(x)\}$ (which  already has probability 1). This is not possible because the probability can not be greater than~1 . Note that we can always choose $y$ with $Sim(x,y) = S_0+\delta > S_0$ and $cS_0 > Sim(x,x)$ $\forall S_0$ and $\forall c < 1$.  This completes the proof.$\hfill\Box$
\end{theorem}

Note that in~\cite{Proc:Charikar} it was shown that we can not have a hash function where the collision probability is equal to the inner product, because ``1 - inner product'' does not satisfy the triangle inequality. This does not totally eliminates the existence of LSH. For instance, under L2Hash, the collision probability is a monotonic function of distance and not the distance itself.

\subsection{Our Proposal: Asymmetric LSH (ALSH)}

The basic idea of LSH is probabilistic bucketing and it is more general than the requirement of having a single hash function $h$. In the LSH algorithm (Section~\ref{sec:fastSimSearch}), we use the same hash function $h$ for both the preprocessing step and the query step. We assign buckets in the hash table to all the candidates  $x \in \mathcal{S}$ using $h$. We use the same $h$ on the query $q$ to identify relevant buckets.  The only requirement for the proof, of Fact 1, to work is that the collision probability of the event $\{h(q) = h(x)\}$  increases with the similarity $Sim(q,x)$. The theory~\cite{Article:LSH_12} behind LSH still works if we use hash function $h_1$ for preprocessing $x \in \mathcal{S}$ and a different hash function $h_2$ for querying, as long as the probability of the event $\{h_2(q) = h_1(x)\}$ increases with $Sim(q,x)$, and there exist $p_1$ and $p_2$ with the required property. The traditional LSH definition does not allow this asymmetry but it is not a required condition in the proof. For this reason, we can relax the definition of $c$-NN without loosing  runtime guarantees.

As the first step, we define a modified locality sensitive hashing, in a slightly different form which will be useful later.\\

\noindent\textbf{Definition:} ({\bf\it Asymmetric} Locality Sensitive Hashing (ALSH))\  A family $\mathcal{H}$, along with the two vector functions  $Q:\mathbb{R}^D \mapsto \mathbb{R}^{D'}$ ({\bf Query Transformation}) and $P:\mathbb{R}^D \mapsto \mathbb{R}^{D'}$ ({\bf Preprocessing Transformation}), is called $(S_0,cS_0,p_1,p_2)$-sensitive if for a given $c$-NN instance with query $q$, and the hash function $h$ chosen uniformly from $\mathcal{H}$ satisfies the following:
\begin{itemize}
\item if $Sim(q,x)\ge S_0$ then $Pr_\mathcal{H}(h(Q(q))) = h(P(x))) \ge p_1$
\item if $ Sim(q,x)\le cS_0$ then $Pr_\mathcal{H}(h(Q(q)) = h(P(x))) \le p_2$
\end{itemize}
Here $x$ is any point in the collection $\mathcal{S}$.\\
%\end{mydef}

When $Q(x) = P(x) = x$, we recover the vanilla LSH definition with $h(.)$ as the required hash function. Coming back to the problem of MIPS, if $Q$ and $P$ are different, the event $\{ h(Q(x)) = h(P(x))\}$ will not have probability equal to 1 in general. Thus, $Q \ne P$ can counter the fact that self similarity is not highest with inner products. We just need the probability of the new collision event $\{ h(Q(q)) = h(P(y))\}$ to satisfy the conditions of Definition of $c$-NN for $Sim(q,y) = q^Ty$.  Note that the query transformation $Q$ is only applied on the query and the pre-processing transformation $P$ is applied to $x \in \mathcal{S}$ while creating hash tables.  It is this asymmetry which will allow us to solve MIPS efficiently. In Section~\ref{sec:construction}, we explicitly show a construction (and hence the existence) of asymmetric locality sensitive hash function for solving MIPS. The source of randomization $h$ for both $q$ and  $x \in \mathcal{S}$ is the same.  \\

Formally, it is not difficult to show a result analogous to Fact 1.
\begin{theorem}\label{theo:extendedLSH}
 Given a family of hash function $\mathcal{H}$ and the associated query and preprocessing transformations $P$ and $Q$, which is $(S_0,cS_0,p_1,p_2)$ -sensitive, one can construct a data structure for $c$-NN with $O(n^\rho \log{n})$ query time and space $O(n^{1 + \rho})$, where $\rho = \frac{\log{p_1}}{\log{p_2}}$.\\

\noindent\textbf{Proof:}\ Use the standard LSH procedure (Section~\ref{sec:fastSimSearch}) with a slight modification. While preprocessing, we assign $x_i$ to bucket $B_l(P(x_i))$ in table $l$.  While querying with query $q$, we retrieve elements from bucket $B_l(Q(q))$ in the hash table $l$. By definition of asymmetric LSH, the probability of retrieving an element, under this modified scheme, follows the same expression as in the original LSH. The proof can be completed by exact same arguments used for proving \textbf{Fact 1} (See ~\cite{Article:LSH_12} for details). $\hfill\Box$
\end{theorem}

\subsection{From MIPS to Near Neighbor Search (NNS)}
\label{sec:construction}
Without loss of generality, we can assume that for the problem of MIPS the query $q$ is normalized, i.e., $||q||_2 = 1$, because in computing $p = \arg\max_{x \in \mathcal{S}} q^Tx$ the argmax is independent of $||q||_2$.  In particular, we can choose to let $U < 1$ be a number such that
\begin{align}
||x_i||_2 \le U < 1,\ \ \forall x_i \in \mathcal{S}
\end{align}
If this is not the case then during the one time preprocessing we can always divide all $x_i$s by $max_{x_i \in \mathcal{S}} \frac{||x_i||_2}{U}$. Note that scaling all $x_i$'s by the same constant does not change $\arg\max_{x \in \mathcal{S}} q^Tx$. 

We are now ready to describe the key step in our algorithm. First, we define two vector transformations $P:\mathbb{R}^D \mapsto \mathbb{R}^{D+m}$ and $Q:\mathbb{R}^D \mapsto  \mathbb{R}^{D+m}$ as follows:
\begin{align}
\label{eq:P}P(x) &= [x; ||x||^2_2;||x||^4_2; ....;||x||^{2^m}_2]\\
\label{eq:Q}Q(x) &= [x; 1/2; 1/2; ....; 1/2],
\end{align}
where [;] is the concatenation. $P(x)$ appends $m$ scalers of the form $||x||_2^{2^i}$ at the end of the vector $x$, while Q(x) simply appends $m$ ``1/2'' to the end of the vector $x$.\\

By observing that
\begin{align}
&||P({x_i})||^2_2 = ||x_i||^2_2 + ||x_i||^4_2+ ... + ||x_i||^{2^m}_2 +  ||x_i||^{2^{m+1}}_2\\
&||Q({q})||^2_2= ||q||^2_2 + m/4 = 1+ m/4\\
&Q(q)^TP({x_i}) = q^Tx_i + \frac{1}{2}( ||x_i||^2_2 + ||x_i||^4_2+ ... + ||x_i||^{2^m}_2)
\end{align}
 we obtain the following key equality:
\begin{equation}\label{eq:errorterminEquation}||Q(q) - P({x_i})||^2_2 = (1 +  m/4) -  2q^Tx_i  +||x_i||^{2^{m+1}}_2\end{equation}
Since $||x_i||_2 \le U < 1$
$$||x_i||^{2^{m+1}}  \rightarrow 0,$$
at the tower rate (exponential to exponential). The term $(1 +  m/4)$ is a fixed constant. As long as $m$ is not too small (e.g., $m\geq3$ would suffice), we have
\begin{equation}
\label{eq:MIPSNNS}
\arg\max_{x \in \mathcal{S}} q^Tx  \simeq \arg\min_{x \in \mathcal{S}} ||Q(q) - P({x})||_2
\end{equation}

To the best of our knowledge, this is the first connection between solving un-normalized MIPS and approximate near neighbor search. Transformations $P$ and $Q$, when norms are less than 1, provide correction to the L2 distance $||Q(q) - P({x_i})||_2$ making it rank correlate with the (un-normalized) inner product. This works only after shrinking the norms, as norms greater than 1 will instead blow the term $||x_i||^{2^{m+1}}_2.$ This interesting mathematical phenomenon connects MIPS with approximate near neighbor search.

\subsection{Fast Algorithms for MIPS}

Eq. (\ref{eq:MIPSNNS}) shows that MIPS reduces to the standard approximate near neighbor search problem which can be efficiently solved. As the error term $||x_i||^{2^{m+1}}_2 < U^{2^{m+1}}$ goes to zero at a tower rate, it quickly becomes negligible for any practical purposes.  In fact, from theoretical perspective, since we are interested in guarantees for $c$-approximate solutions, this additional error can be absorbed in the approximation parameter $c$. \\

Formally, we can state the following theorem.
\begin{theorem}
\label{theo:collprobnew}
Given a $c$-approximate  instance of MIPS, i.e., $Sim(q,x) = q^Tx$, and a query $q$ such that $||q||_2 = 1$ along with a collection $\mathcal{S}$ having $||x||_2 \le  U <1$  $\forall x \in \mathcal{S}.$ Let $P$ and $Q$ be the vector transformations defined in Eq. (\ref{eq:P}) and Eq. (\ref{eq:Q}), respectively.  We have the following two conditions for hash function $h_{a,b}^{L2}$ (defined by Eq. (\ref{eq:L2Hash}))
\begin{itemize}
\item if $q^Tx \ge S_0$ then $$\hspace{-0.3in}Pr[h_{a,b}^{L2}(Q(q)) = h_{a,b}^{L2}(P(x))] \ge F_r\big(\sqrt{1 + m/4 - 2S_0 + U^{2^{m+1}}}\big)$$
\item if $ q^Tx \le cS_0$ then
\begin{align}\notag
\hspace{-0.3in}Pr[h_{a,b}^{L2}(Q(q)) = h_{a,b}^{L2}(P(x))] \le F_r\big(\sqrt{1+m/4 - 2cS_0}\big)
\end{align}
where the function $F_r$ is defined by Eq. (\ref{eq:F}).%: $F(d) =  1 - 2\Phi(-r/d) - \frac{2}{\sqrt{2\pi}(r/d)}\left(1 - e^{-(r/d)^2/2}\right)$.

\end{itemize}
\noindent\textbf{Proof:}\ \
From Eq. (\ref{eq:L2collprob}), we have
\begin{align}\notag
 &Pr[h_{a,b}^{L2}(Q(q)) = h_{a,b}^{L2}(P(x))]\\\notag
=& F_r(||Q(q) - P(x)||_2) \\\notag
=&  F_r\big(\sqrt{1 + m/4 - 2q^Tx + ||x||_2^{2^{m+1}}}\big) \\\notag
 \ge& F_r\big(\sqrt{1 + m/4 - 2S_0 + U^{2^{m+1}}}\big)
\end{align}
The last step follows from the monotonically decreasing nature of $F$ combined with inequalities $q^Tx \ge S_0$ and $||x||_2 \le  U$. We have also used the monotonicity of the square root function. The second inequality similarly follows using  $q^Tx \le cS_0$ and $||x||_2 \ge  0.$. This completes the proof.$\hfill\Box$\\

\end{theorem}

The conditions  $||q||_2 = 1$ and $||x||_2 \le U < 1,\ \forall  x \in \mathcal{S}$ can be absorbed in the transformations $Q$ and $P$ respectively, but we show it explicitly for clarity.

Thus, we have obtained $p_1 =  F_r\big(\sqrt{(1 + m/4) - 2S_0 + U^{2^{m+1}}}\big)$ and $p_2 = F_r\big(\sqrt{(1+m/4) - 2cS_0}\big)$. Applying Theorem~\ref{theo:extendedLSH}, we can construct data structures with  worst case $O(n^\rho \log{n})$ query time guarantees for $c$-approximate MIPS, where
\begin{align}\rho = \frac{\log{F_r\big(\sqrt{1 + m/4- 2S_0 + U^{2^{m+1}}}\big)}}{\log{ F_r\big(\sqrt{1+m/4 - 2cS_0\big)}}}
 \end{align}
 We also need $p_1 > p_2$ in order for $\rho < 1$. This requires us to have $- 2S_0 + U^{2^{m+1}} < -2cS_0$, which boils down to the condition $c < 1 - \frac{U^{2^{m+1}}}{2S_0}$. Note that $\frac{U^{2^{m+1}}}{2S_0}$ can be made arbitrarily close to zero with the appropriate value of $m$. For any given $c <1$, there always exist $U <1$ and $m$ such that $\rho < 1$.  This way, we obtain a sublinear query time algorithm for MIPS. The guarantee holds for any values of $U$ and $m$ satisfying $m \in \mathbb{N}^{+}$ and $U \in (0,1)$. We also have one more parameter $r$ for the hash function $h_{a,b}$. Recall the definition of $F_r$ in Eq. (\ref{eq:F}): $F_r(d) =  1 - 2\Phi(-r/d) - \frac{2}{\sqrt{2\pi}(r/d)}\left(1 - e^{-(r/d)^2/2}\right)$.\\

\begin{figure}[h!]
\begin{center}
\mbox{
\includegraphics[width=2.7in]{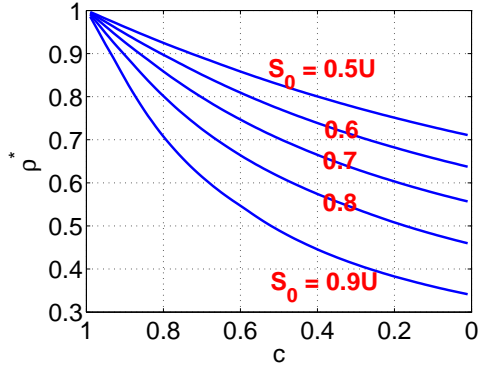}}
\end{center}
\vspace{-0.2in}
\caption{Optimal values of $\rho^*$ with respect to approximation ratio $c$ for different $S_0$. The optimization of Eq. (\ref{eq:optrho}) was done by a grid search over parameters $r$, $U$ and $m$, given $S_0$ and $c$. See Figure~\ref{fig:Optchoices} for the corresponding optimal values of parameters.}\label{fig:OptRho}
\end{figure}

 Given a $c$-approximate MIPS instance, $\rho$ is a function of 3 parameters: $U$, $m$, $r$. The algorithm with the best query time chooses $U$, $m$ and $r$, which minimizes the value of $\rho$. For convenience, we define
\begin{align}
\label{eq:optrho}
\rho^* = \min_{U,m,r} \frac{\log{F_r\big(\sqrt{1 + m/4- 2S_0 + U^{2^{m+1}}}\big)}}{\log{ F_r\big(\sqrt{1+m/4 - 2cS_0}\big)}}\\
\nonumber s.t. \hspace{0.1in} \frac{U^{2^{m+1}}}{2S_0} <  1 -  c, \hspace{0.05in}  m \in \mathbb{N}^{+}, \hspace{0.05in} r > 0 \mbox{ and } 0 < U < 1.
\end{align}
See Figure~\ref{fig:OptRho} for the plots of $\rho^*$. With this best value of $\rho$, we can state our main result in Theorem~\ref{theo:main}.
\newpage

\begin{theorem}
\label{theo:main}
({\bf Approximate MIPS is Efficient})  For the problem of $c$-approximate MIPS, one can construct a data structure having $O(n^{\rho^*} \log{n})$ query time and space $O(n^{1+\rho^*})$, where $\rho^* < 1$  is the solution to constraint optimization (\ref{eq:optrho}).
\end{theorem}

Just like in the typical LSH framework, the value of $\rho^*$ in Theorem~\ref{theo:main} depends on the $c$-approximate instance we aim to solve, which requires knowing the similarity threshold $S_0$ and the approximation ratio $c$. Since, $||q||_2 = 1$ and $||x||_2 \le U < 1,\  \forall  x \in \mathcal{S}$, we have $q^tx \le U$. A reasonable choice of the threshold $S_0$ is to choose a high fraction of U, for example,  $S_0 = 0.9U$ or $S_0 = 0.8U$.\\

\begin{figure}[h!]
\begin{center}
\mbox{
\includegraphics[width=2.3in]{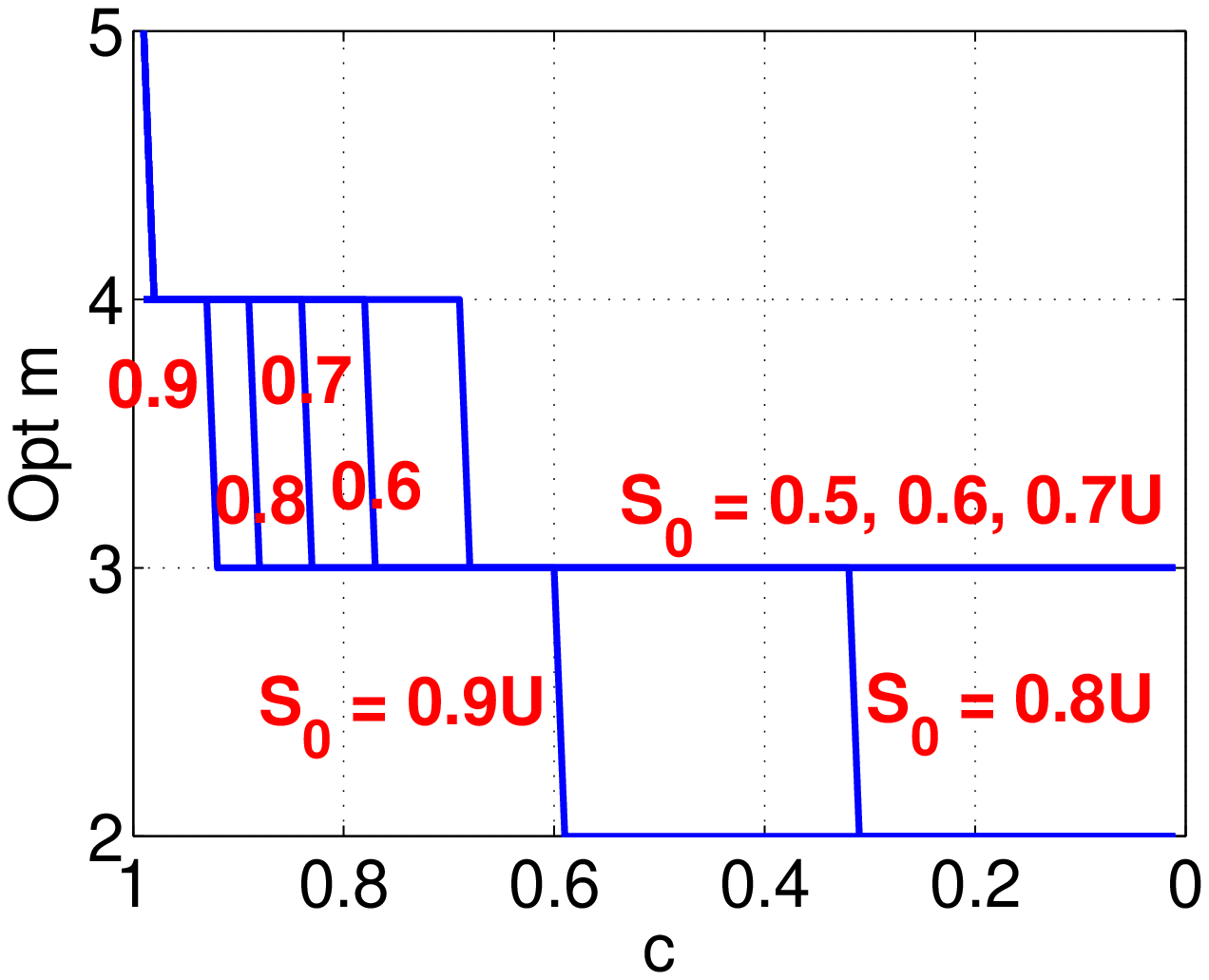}\hspace{-0.15in}
\includegraphics[width=2.3in]{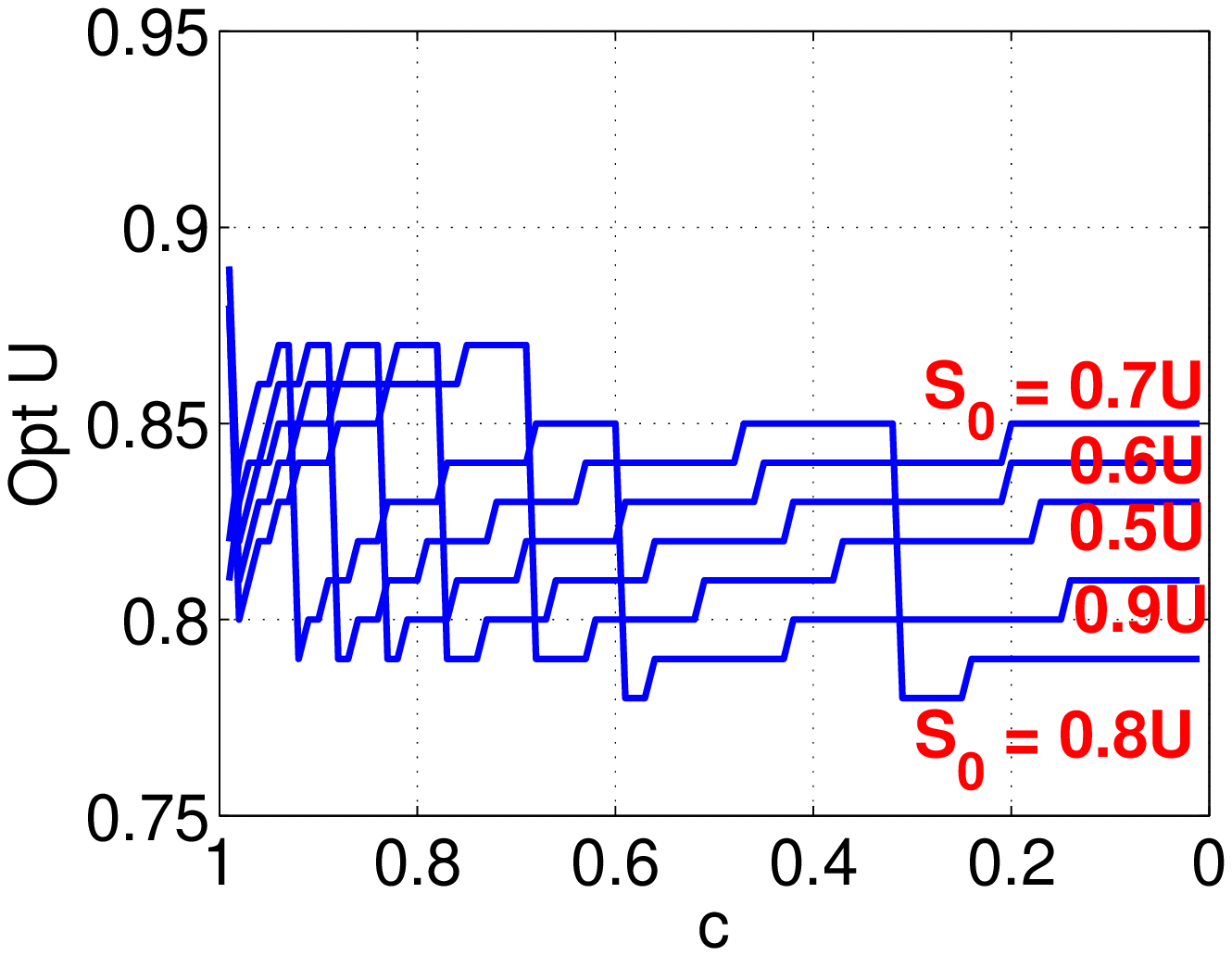}\hspace{-0.15in}
\includegraphics[width=2.3in]{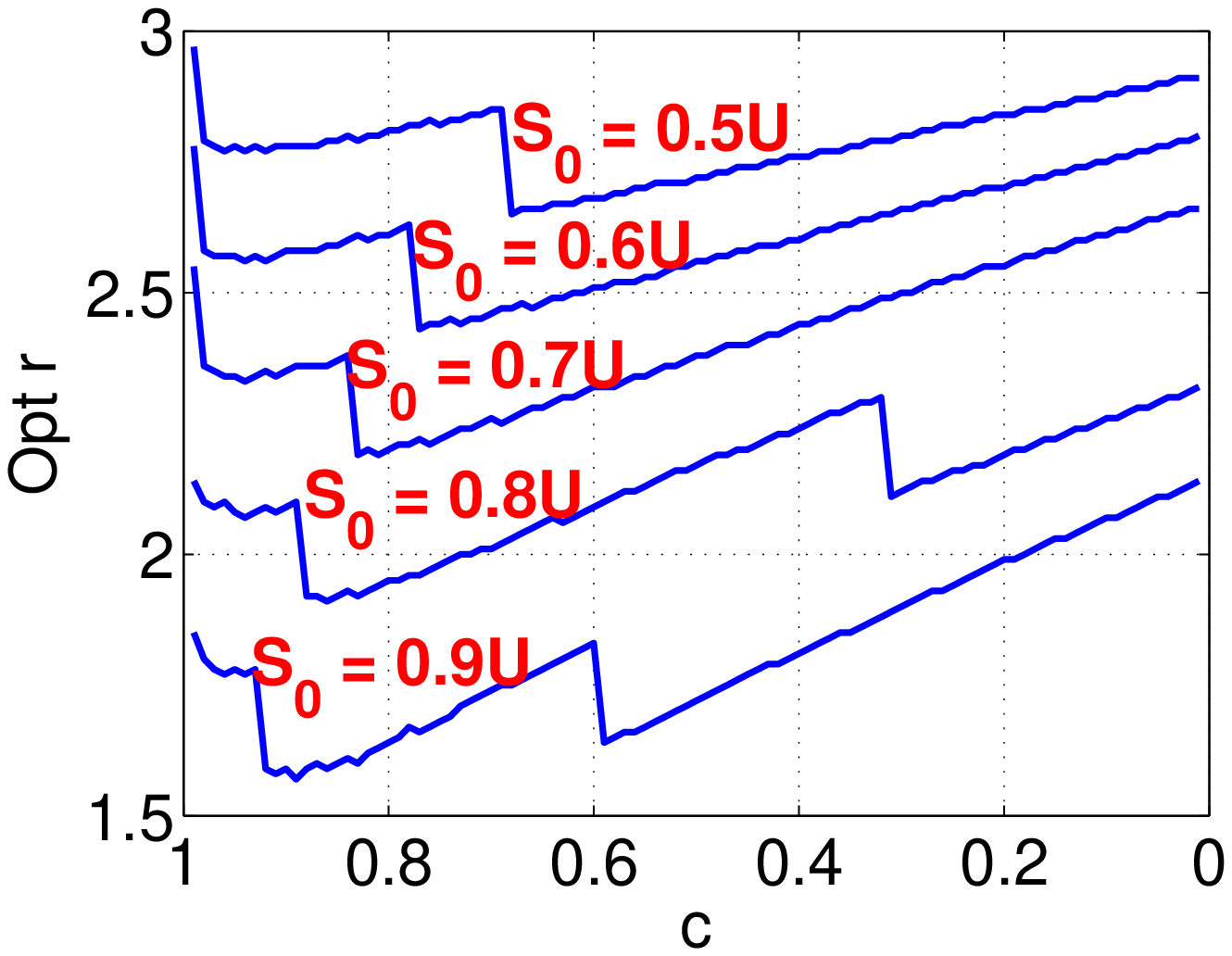}}
\end{center}
\vspace{-0.2in}
\caption{Optimal values of parameters $m$, $U$ and $r$ with respect to the approximation ratio $c$ for relatively high similarity threshold $S_0$. The corresponding optimal values of $\rho^*$ are shown in Figure~\ref{fig:OptRho}.}\label{fig:Optchoices}
\end{figure}

The computation of $\rho^*$ and the optimal values of corresponding parameters can be conducted via a grid search over the possible values of $U$, $m$ and $r$, as we only have  3 parameters.  We compute the values of $\rho^*$ along with the corresponding optimal values of  $U$, $m$ and $r$ for $S_0 \in \{0.9U,\ 0.8U,\ 0.7U,\ 0.6U,\ 0.5U\}$ for different approximation ratios $c$ ranging from 0 to 1. The plot of the optimal $\rho^*$ is shown in  Figure~\ref{fig:OptRho}, and the corresponding  optimal values of $U$, $m$ and $r$ are shown in Figure~\ref{fig:Optchoices}.

\subsection{Practical Recommendation of Parameters}

\begin{figure}[h!]
\begin{center}
\mbox{
\includegraphics[width=2.7in]{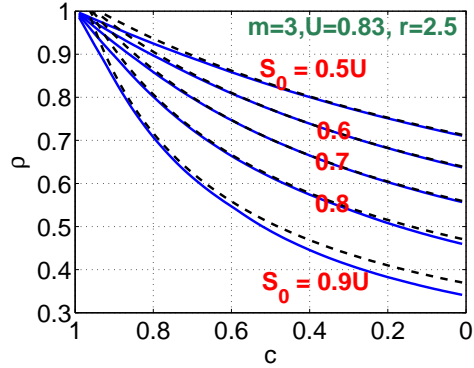}
}
\end{center}
\vspace{-0.3in}
\caption{$\rho$ values (dashed curves) for $m=3$, $U=0.83$ and $r=2.5$. The solid curves are the optimal $\rho^*$ values as shown in Figure~\ref{fig:OptRho}. }\label{fig:Rhom3}
\end{figure}

In practice, the actual choice of $S_0$ and $c$ is dependent on the data and is often unknown.  Figure~\ref{fig:Optchoices} illustrates that $m \in\{2, 3, 4\}$, $U \in [0.8,\ 0.85]$, and $r \in [ 1.5,\ 3]$  are reasonable choices.  For convenience, we recommend to use $m = 3$, $U = 0.83$, and $r = 2.5$. With this choice of the parameters,  Figure~\ref{fig:Rhom3} shows that the $\rho$ values using these parameters are very close to the optimal $\rho^*$ values.

\vspace{-0.2in}
\subsection{More Insight: The Trade-off between $U$ and $m$}\label{sec:recomend}

Let $\epsilon =  U^{2^{m+1}}$ be the error term in Eq. (\ref{eq:errorterminEquation}). As long as $\epsilon$  is small the MIPS problem reduces to standard near neighbor search via the transformations $P$ and $Q$.  There are two ways to make $\epsilon$ small, either we choose a small value of $U$ or a large value of $m$. There is an interesting trade-off between parameters $U$ and $m$.  To see this, consider the following  Figure~\ref{fig:Collprob} for $F_r$(d), i.e., the collision probability defined in Eq.~(\ref{eq:F}).

\begin{figure}[h!]
\begin{center}
\mbox{
\includegraphics[width=2.7in]{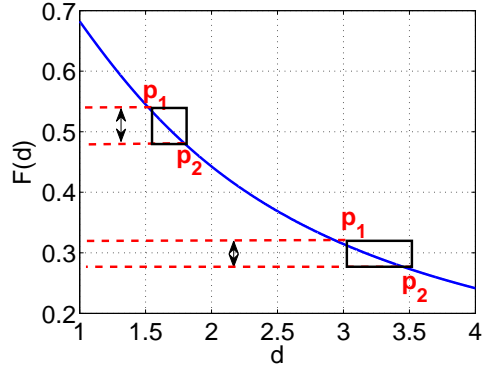}}
\end{center}
\vspace{-0.3in}
\caption{Plot of the monotonically decreasing function $F_r(d)$  (Eq. (\ref{eq:F})). The curve is steeper when $d$ is close to 1, and flatter when $d$ goes away from 1. This leads to a tradeoff between the choices of $U$ and $m$. }\label{fig:Collprob}
\end{figure}

Suppose $\epsilon =  U^{2^{m+1}}$ is small, then $p_1 = F_r(\sqrt{1 + m/4 - 2q^Tx})$ and $p_2 = F_r(\sqrt{1 + m/4 - 2cq^Tx})$. Because of the bounded norms, we have $-||q||_2||x||_2 = -U \le q^Tx \le U =||q||_2||x||_2$. Consider the high similarity range $$d \in [\sqrt{1 + m/4 - 2U}, \sqrt{1 + m/4 - 2cU}],$$ for some appropriately chosen $c < 1$. This range is wider, if $U$ is close to 1, if $U$ is close to 0 then  $ 2U \simeq 2cU$  making the arguments to $F$ close to each other, which in turn decreases the gap between $p_1$ and $p_2$.  On the other hand, when $m$ is larger, it adds bigger offset (term $m/4$ inside square root) to both the arguments. Adding offset, makes $p_1$ and $p_2$ shift towards right in the collision probability plot. Right shift, makes $p_1$ and $p_2$ closer because of the nature of the collision probability curve (see Figure~\ref{fig:Collprob}). Thus, bigger $m$ makes $p_1$ and $p_2$ closer. Smaller $m$ pushes $p_1$ and $p_2$ towards left and thereby increasing the gap between them.  Small $\rho =\frac{\log{p_1}}{\log{p_2}}$ roughly boils down to having a bigger gap between $p_1$ and $p_2$~\cite{Book:Raj_Ullman}. Ideally, for small $\rho$, we would like to use big $U$ and small $m$. The error $\epsilon = U^{2^{m+1}}$ exactly follows the opposite trend. For error to be small we want small $U$ and not too small $m$.

\subsection{Parallelization}

To conclude this section,  we should mention that the hash table based scheme is massively parallelizable. Different nodes on cluster need to maintain their own hash tables and hash functions. The operation of retrieving from buckets and computing the maximum inner product over those retrieved candidates, given a query, is a local operation. Computing the final maximum can be conducted efficiently by simply communicating one single number per nodes. Scalability of hashing based methods is one of the  reasons which account for their  popularity in industrial practice.

\section{Evaluations}
\label{sec:evaluations}

\subsection{Datasets}
We evaluate our proposed hash function for the MIPS  problem on  two popular collaborative filtering datasets (on the task of item recommendations):
\begin{itemize}
\item {\bf Movielens.}\ We choose the largest available Movielens dataset, the {\em Movielens 10M}, which contains around 10 million movie ratings from 70,000 users over 10,000 movie titles.  The ratings are between 1 to 5, with increments of 0.5 (i.e.,  10 possible ratings in total).
\item {\bf Netflix.}\ This dataset contains 100 million movie ratings from 480,000 users over 17,000 movie titles. The ratings are on a scale from 1 to 5 (integer).
\end{itemize}
Each dataset forms a sparse \textbf{user-item matrix} $R$, where the value of $R(i,j)$ indicates the rating of user $i$ for movie $j$. Given the user-item ratings matrix $R$, we follow the PureSVD procedure described in~\cite{Proc:Cremonesi_RecSys} to generate user and item latent vectors.  That is, we compute the SVD of the ratings matrix
$$R = W\Sigma V^T$$
where $W$ is $n_{users} \times f$ matrix and $V$ is $n_{item} \times f$ matrix for some appropriately chosen rank $f$ (which is also called the {\em latent dimension}).

After the SVD step, the rows of matrix $U = W\Sigma$ are treated as the user characteristic vectors while rows of matrix $V$ correspond to the item characteristic vectors. More specifically $u_i$, the $i^{th}$ row of matrix $U$, denotes the characteristic vector of user $i$, while the $j^{th}$ row of $V$, i.e., $v_j$, corresponds to the characteristic vector for item  $j$.  The compatibility between item $i$ and item $j$  is given by the inner product between the corresponding user and item characteristic vectors. Therefore, in recommending top-ranked items to users $i$, the PureSVD method returns top-ranked items based on the inner products $u_i^Tv_j,\ \forall j$.

The PureSVD procedure, despite its simplicity, outperforms other popular recommendation algorithms for the task of top-ranking recommendations (see~\cite{Proc:Cremonesi_RecSys} for more details) on these two datasets. Following~\cite{Proc:Cremonesi_RecSys}, we use the same choices for the latent dimension  $f$, i.e., $f=150$ for Movielens and $f=300$ for Netflix.

\subsection{Baseline Hash Function}

Our proposal is the first provable hashing scheme in the literature for retrieving inner products and hence there is no existing baseline. Since, our hash function uses Hashing for L2 distance after asymmetric transformation $P$ (\ref{eq:P}) and $Q$ (\ref{eq:Q}), we would  like to know if such transformations are even needed and furthermore get an estimate of the improvements obtained using these transformations. We therefore compare our proposal with \emph{\bf L2LSH} the hashing scheme $h_{a,b}^{L2}$ described by Eq. (\ref{eq:L2Hash}). It is implemented in the LSH package for near neighbor search with Euclidean distance.

Although, L2LSH are not optimal for retrieving un-normalized inner products, it does  provide some indexing capability. Our experiments will show that the proposed method outperform L2LSH, often significantly so, in retrieving inner products. This is not surprising as we know that the rankings of L2 distance can be different from rankings of inner products.

\subsection{Evaluations}\label{sec:hashquality}

We are interested in knowing, how the two hash functions correlate with the top-$T$ inner products.  For this task, given a user $i$ and its corresponding user vector $u_i$, we compute the top-$T$ gold standard items based on the actual inner products $u_i^Tv_j$, $\forall j$.  We then compute $K$ different hash codes of the vector $u_i$ and all the item vectors $v_j$s.  For every item $v_j$, we then compute the number of times its hash values matches (or collides) with the hash values of query which is user $u_i$, i.e., we compute
\begin{equation}
Matches_j = \sum_{t=1}^{K} \mathbbm{1}(h_t(u_i) = h_t(v_j)),
\end{equation}
where $\mathbbm{1}$ is the indicator function. Based on $Matches_j$ we rank all the items. This procedure generates a sorted list of all the items for a given user vector $u_i$ corresponding to every hash function under consideration.  Here, we use $h = h_{a,b}^{L2}$ for L2LSH. For our proposed asymmetric  hash function we have $h(u_i) = h_{a,b}^{L2}(Q(u_i))$, since $u_i$ is the query, and  $h(v_j) = h_{a,b}^{L2}(P(v_j))$ for all the items. The subscript $t$ is used to distinguish independent draws of $h$. Ideally, for a better hashing scheme, $Matches_j$ should be higher for items having higher inner products with the given user $u_i$.

We compute the precision and  recall of the top-$T$ items for $T\in\{1, 5, 10\}$, obtained from the sorted list based on $Matches$.  To compute this precision and recall, we start at the top of the ranked item list and walk down in order. Suppose we are at the $k^{th}$ ranked item, we check if this item belongs to the gold standard top-$T$ list. If it is one of the top-$T$ gold standard item, then we increment the count of \emph{relevant seen} by 1, else we move to $k+1$. By $k^{th}$ step, we have already seen $k$ items, so the \emph{total items seen} is $k$. The precision and recall at that point is then computed as:
\begin{align}
 Precision = \frac{\text{relevant seen}}{k}, \hspace{0.3in}
Recall = \frac{\text{relevant seen}}{T}
\end{align}
We vary a large number of $k$ values to obtain continuously-looking {\em precision-recall} curves. Note that it is important to balance both precision and recall. Methodology which obtains higher precision at a given recall is superior. Higher precision indicates higher ranking of the relevant items. We  average this value of precision and recall over 2000 randomly chosen users.  \\

\begin{figure}[h!]
\begin{center}
\mbox{
\includegraphics[width=2.25in]{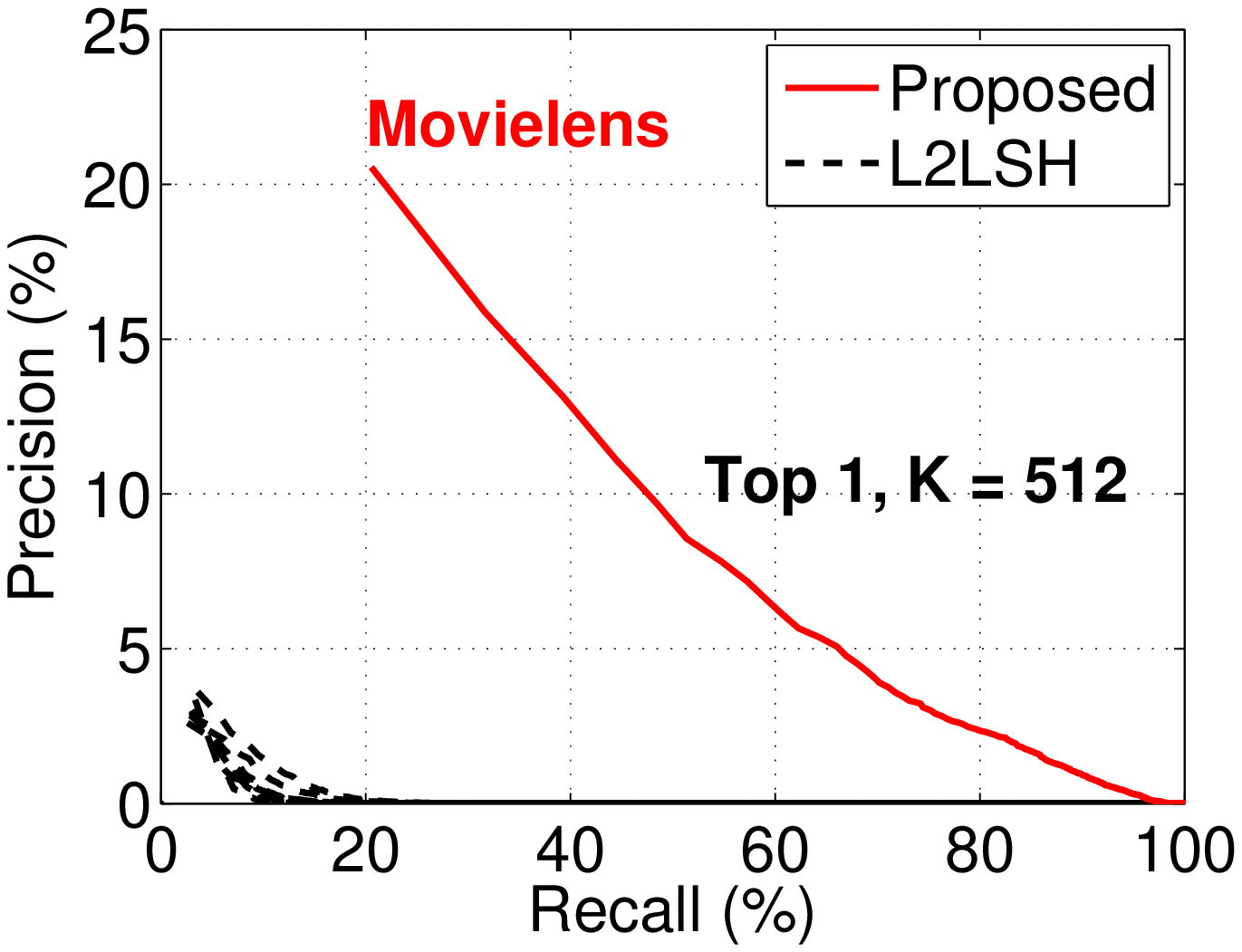}\hspace{-0.13in}
\includegraphics[width=2.25in]{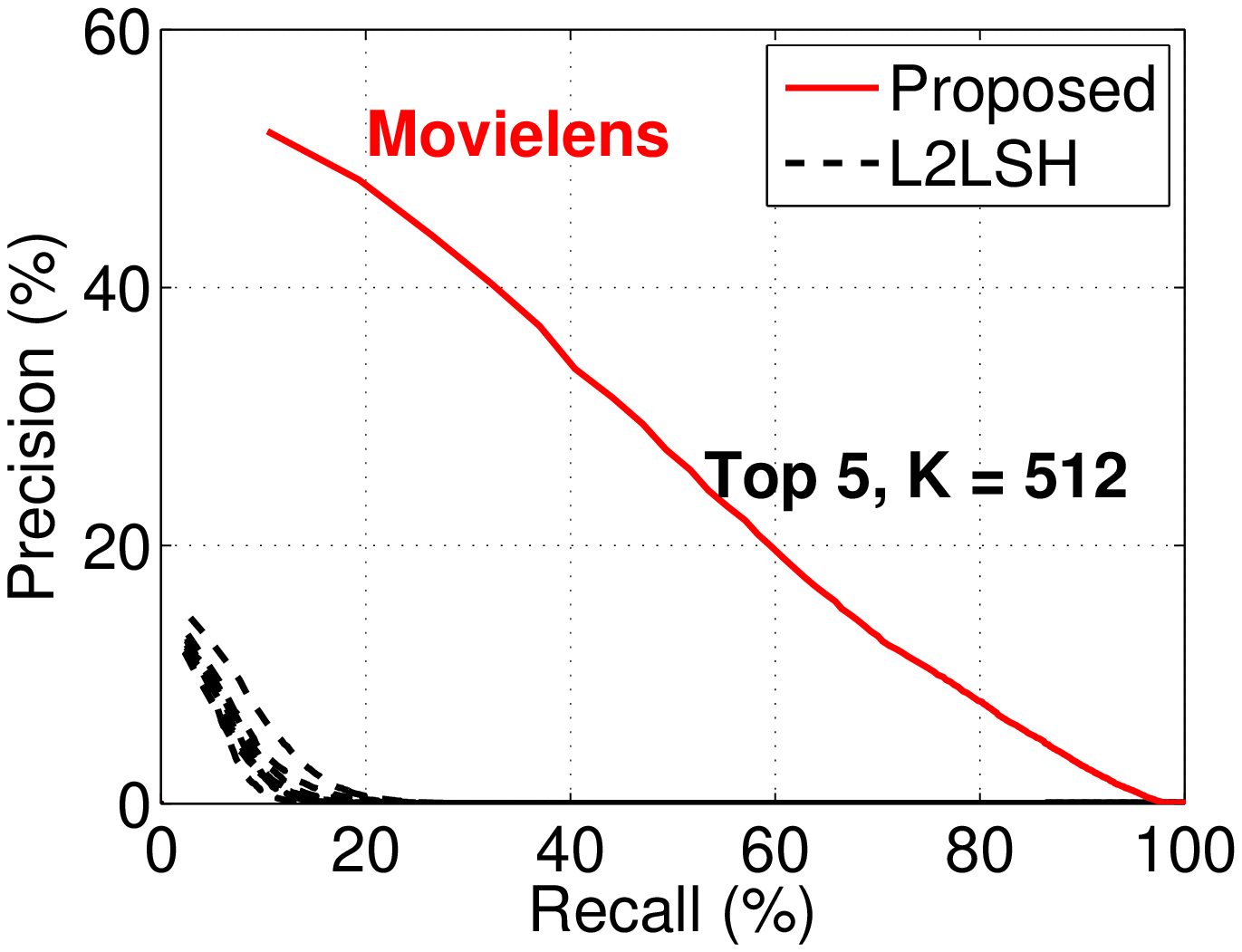}\hspace{-0.13in}
\includegraphics[width=2.25in]{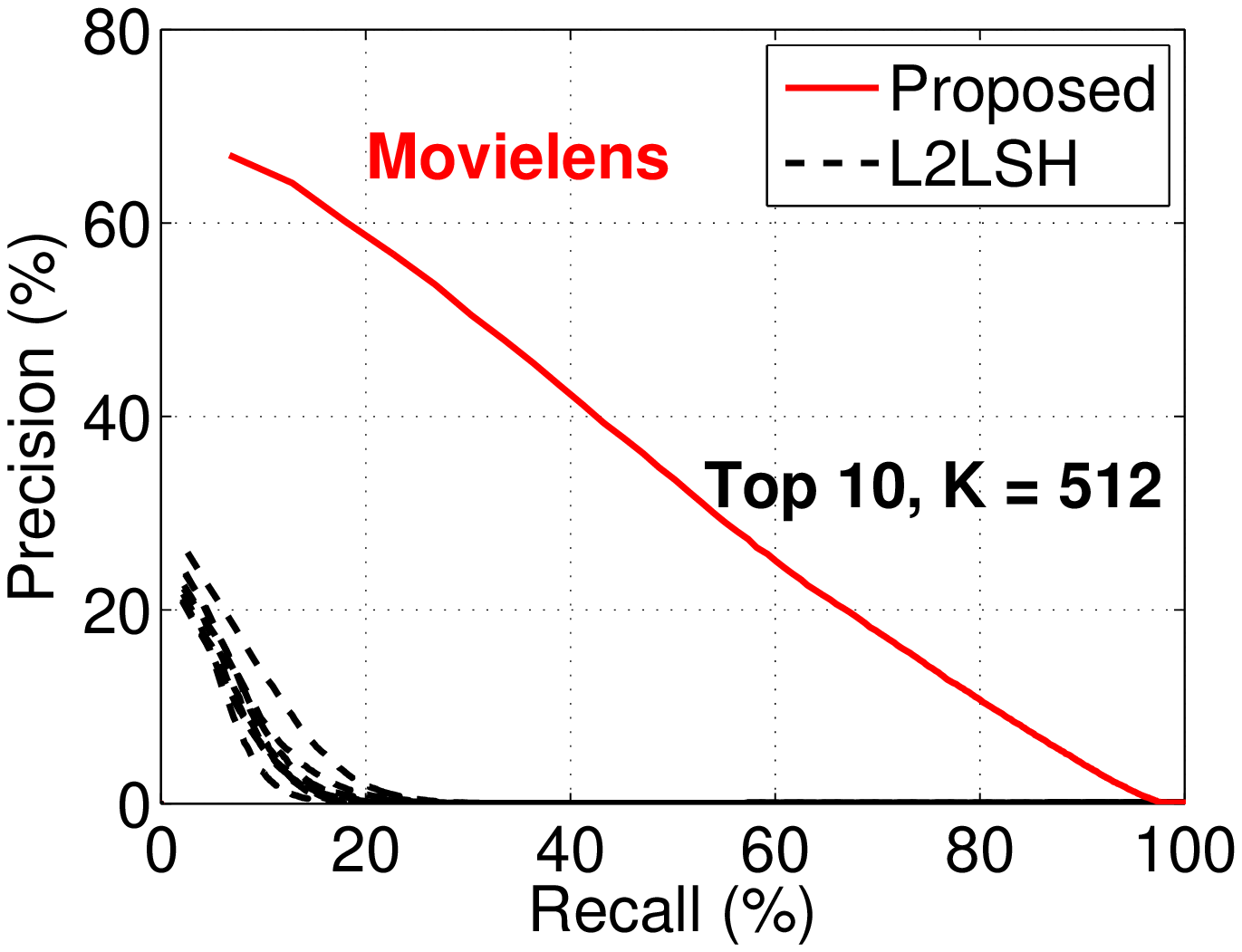}
}

\mbox{
\includegraphics[width=2.25in]{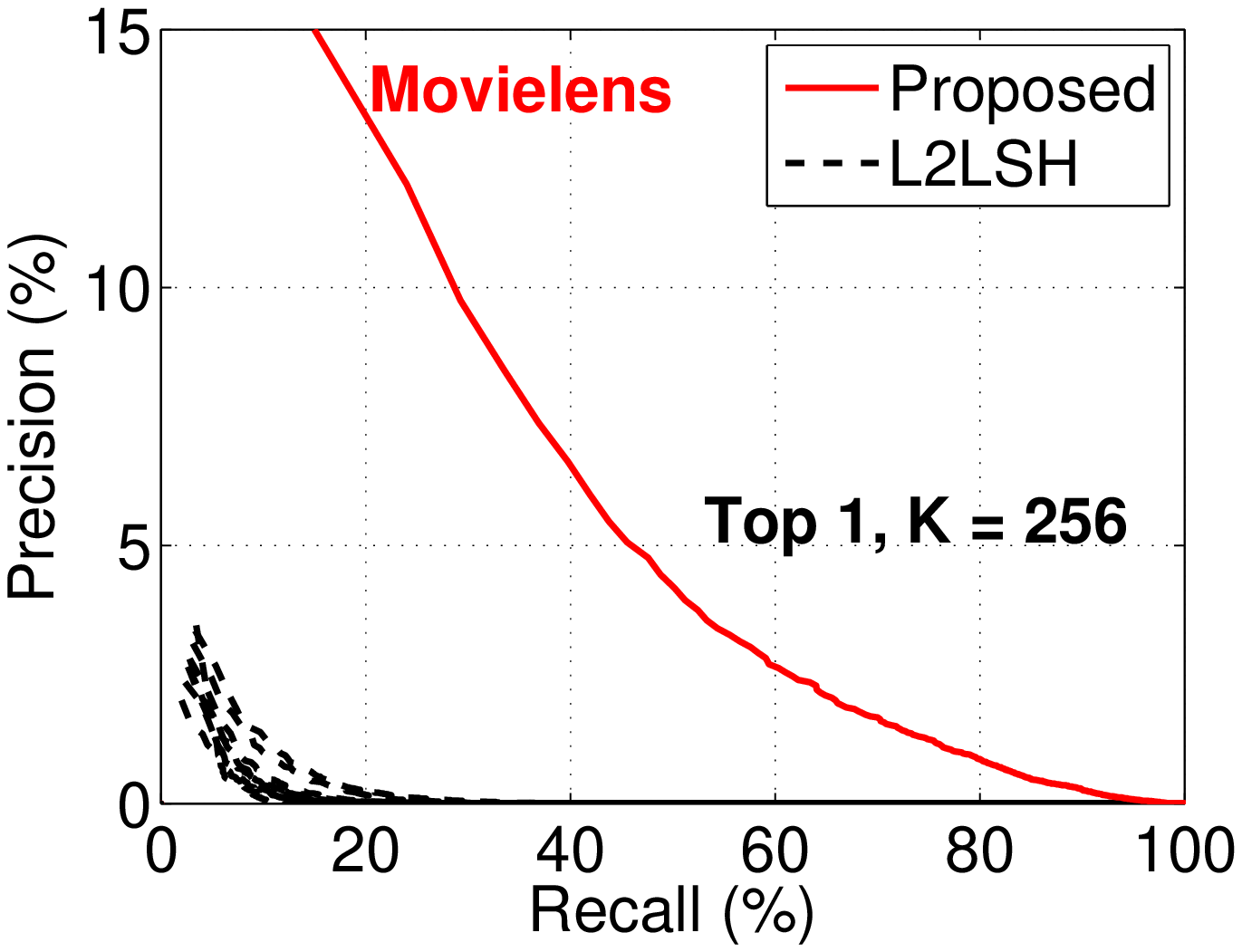}\hspace{-0.13in}
\includegraphics[width=2.25in]{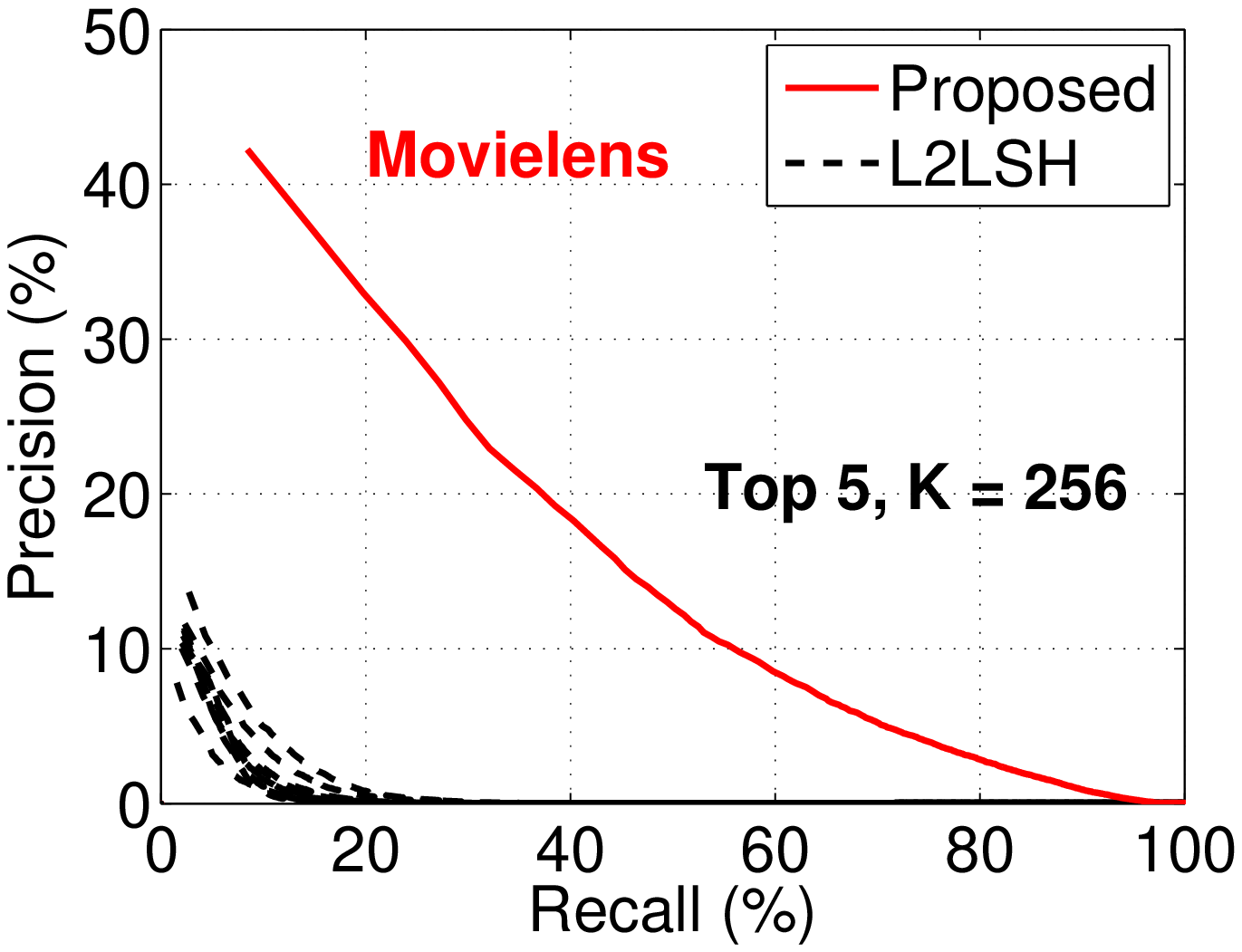}\hspace{-0.13in}
\includegraphics[width=2.25in]{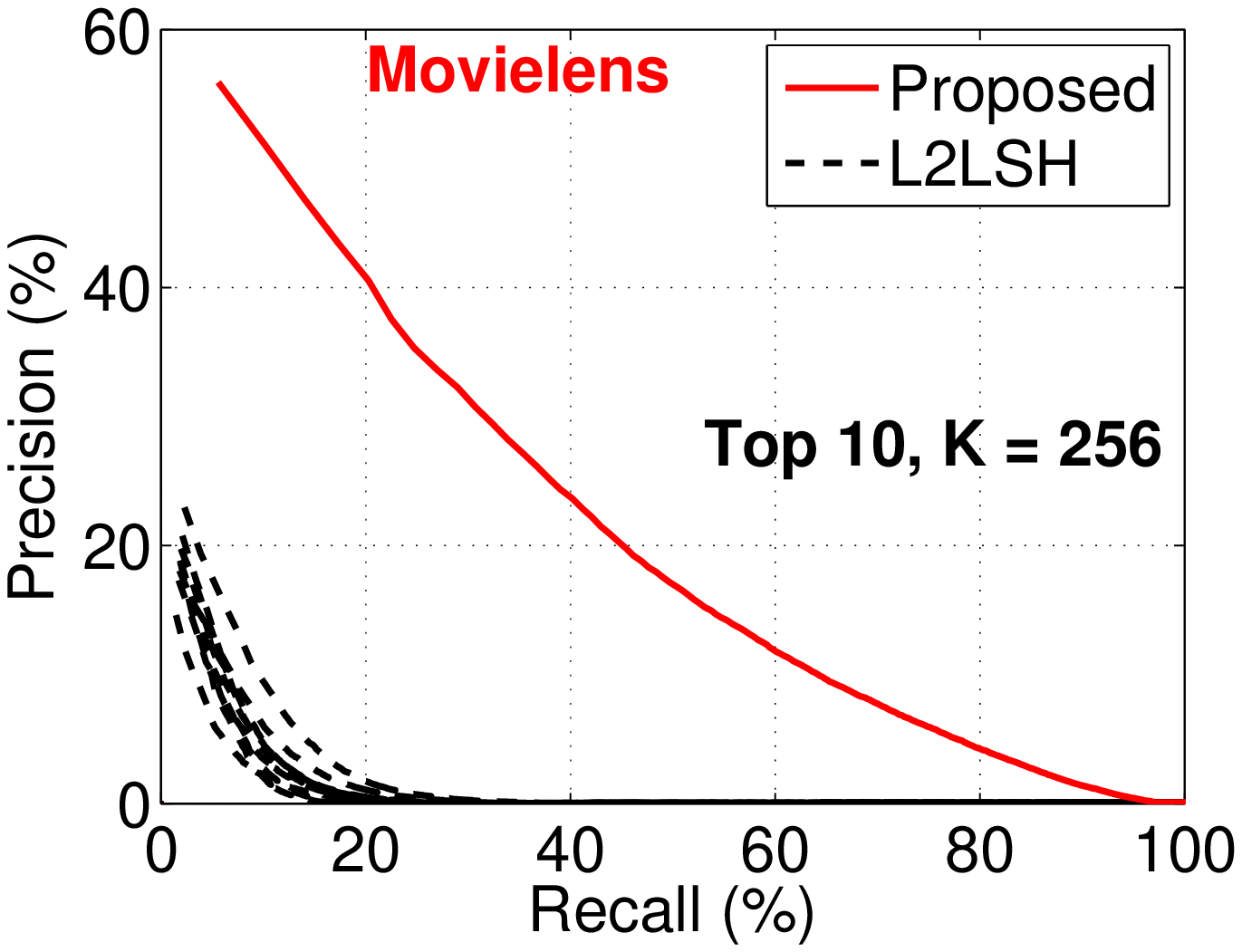}
}

\mbox{
\includegraphics[width=2.25in]{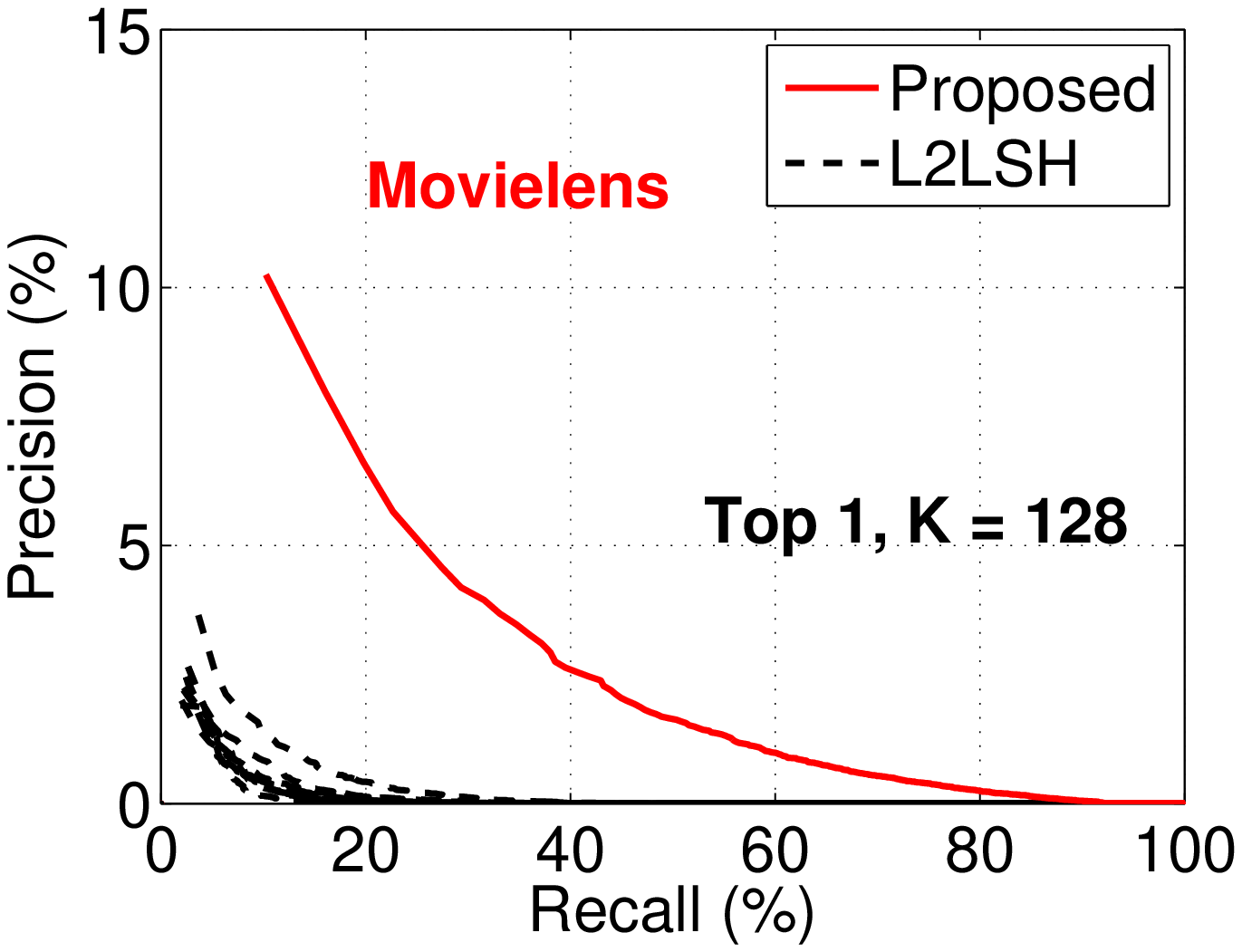}\hspace{-0.13in}
\includegraphics[width=2.25in]{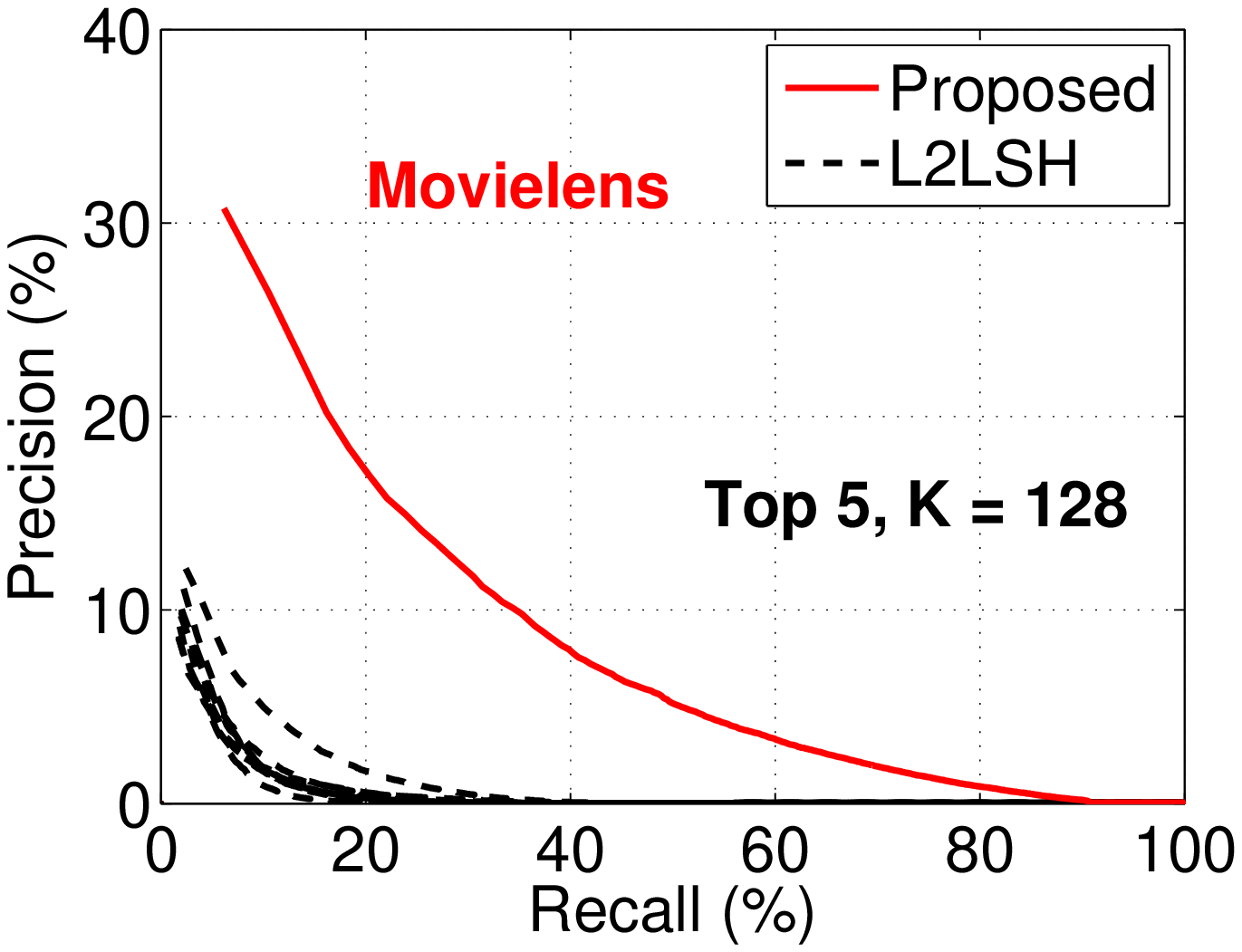}\hspace{-0.13in}
\includegraphics[width=2.25in]{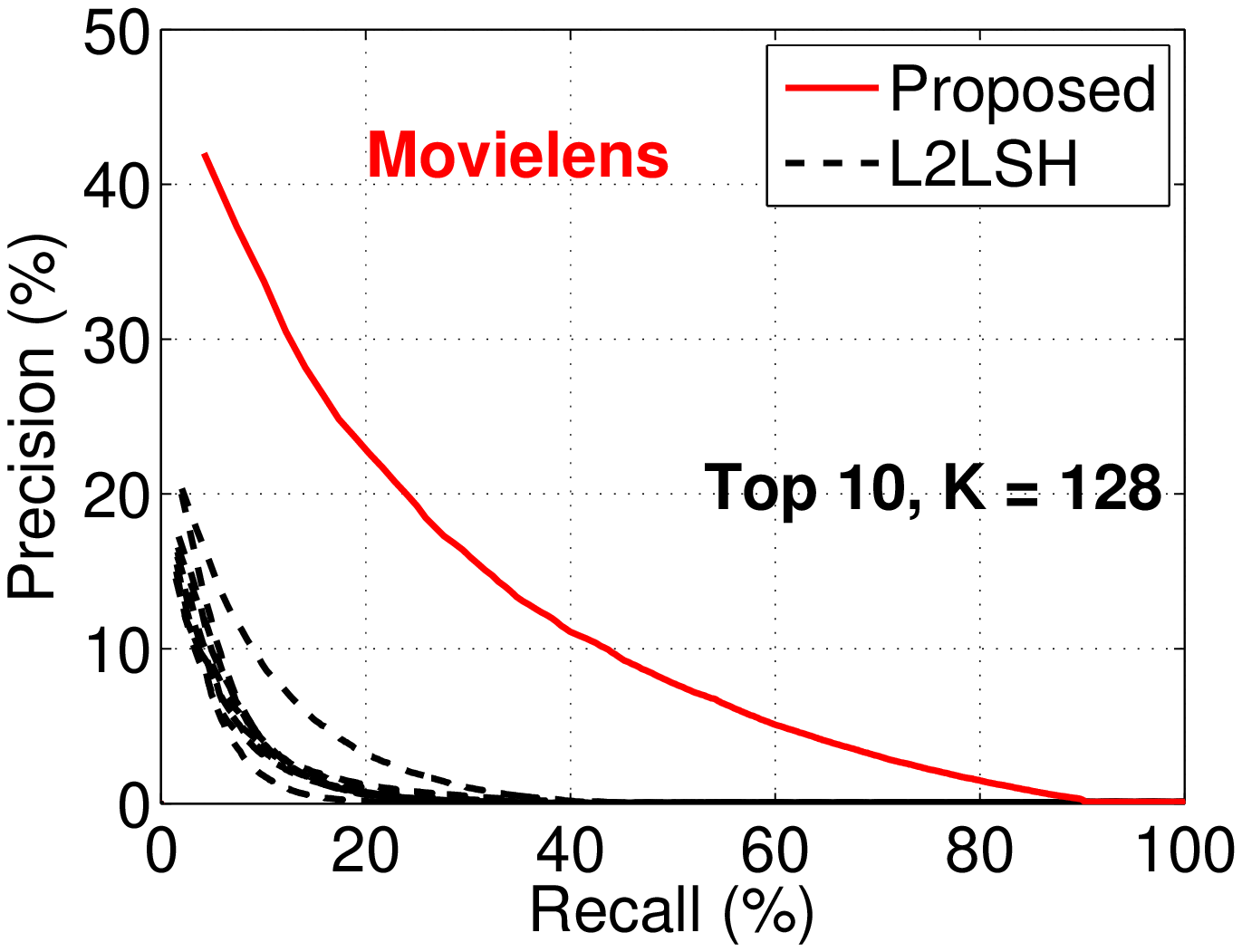}
}

\mbox{
\includegraphics[width=2.25in]{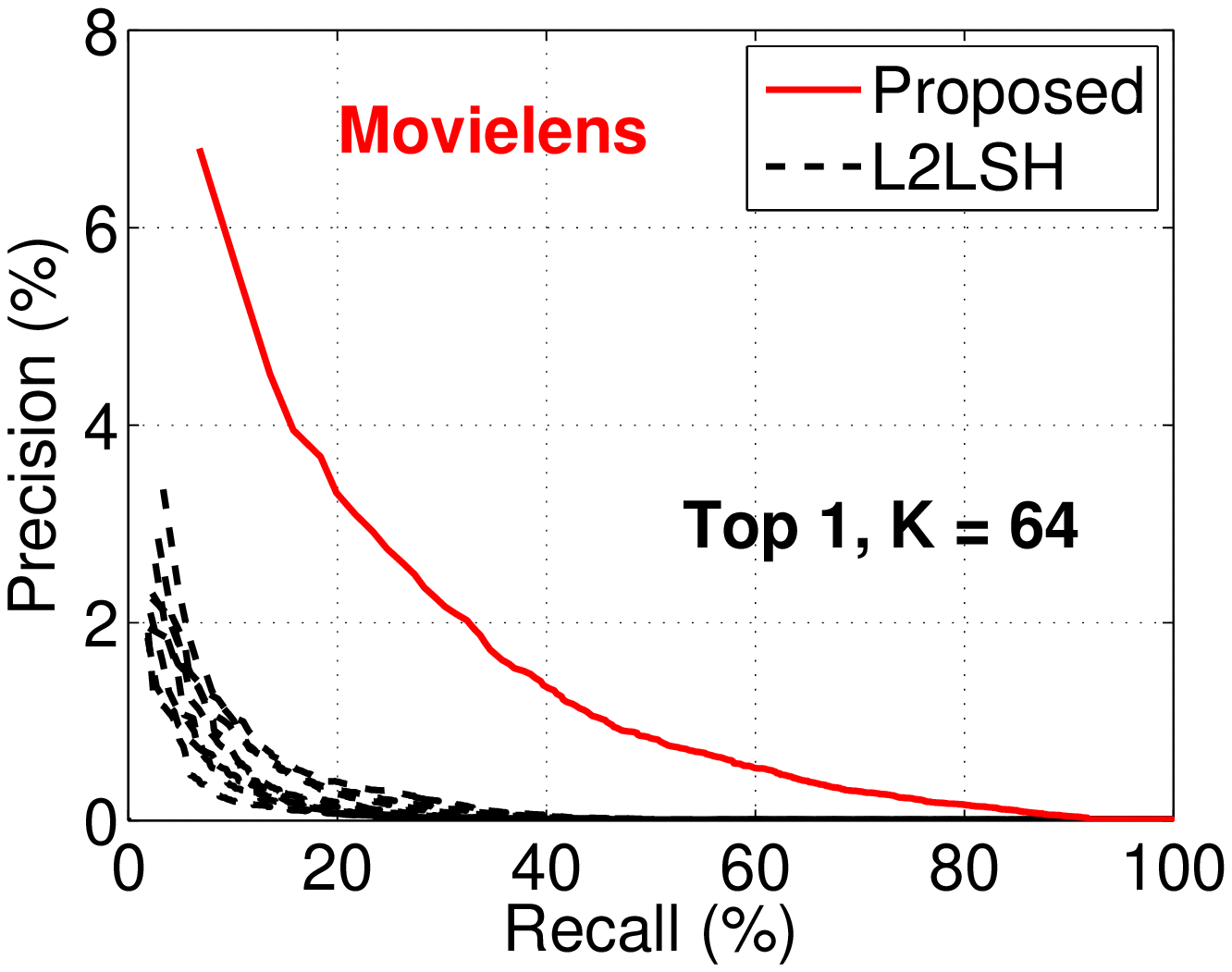}\hspace{-0.13in}
\includegraphics[width=2.25in]{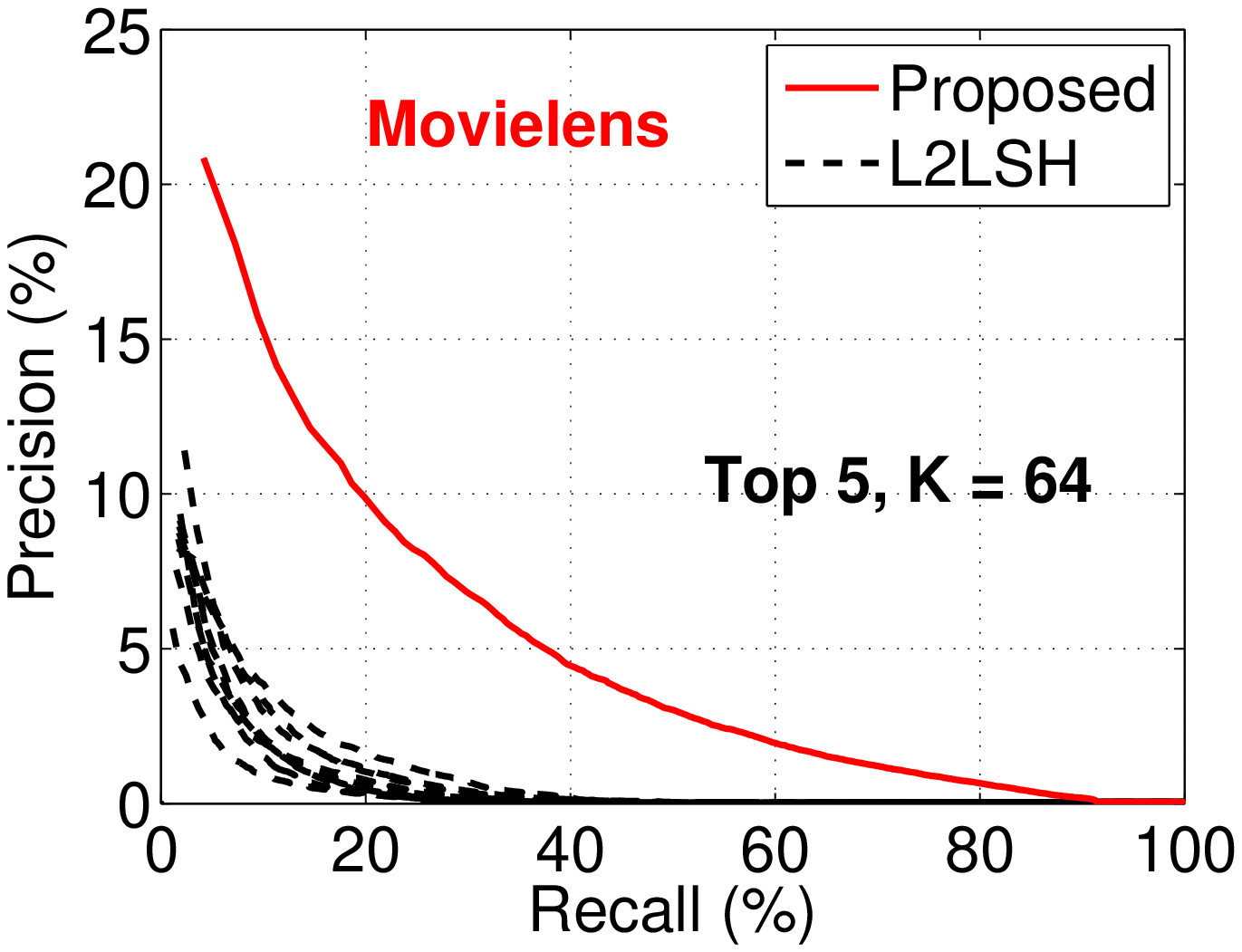}\hspace{-0.13in}
\includegraphics[width=2.25in]{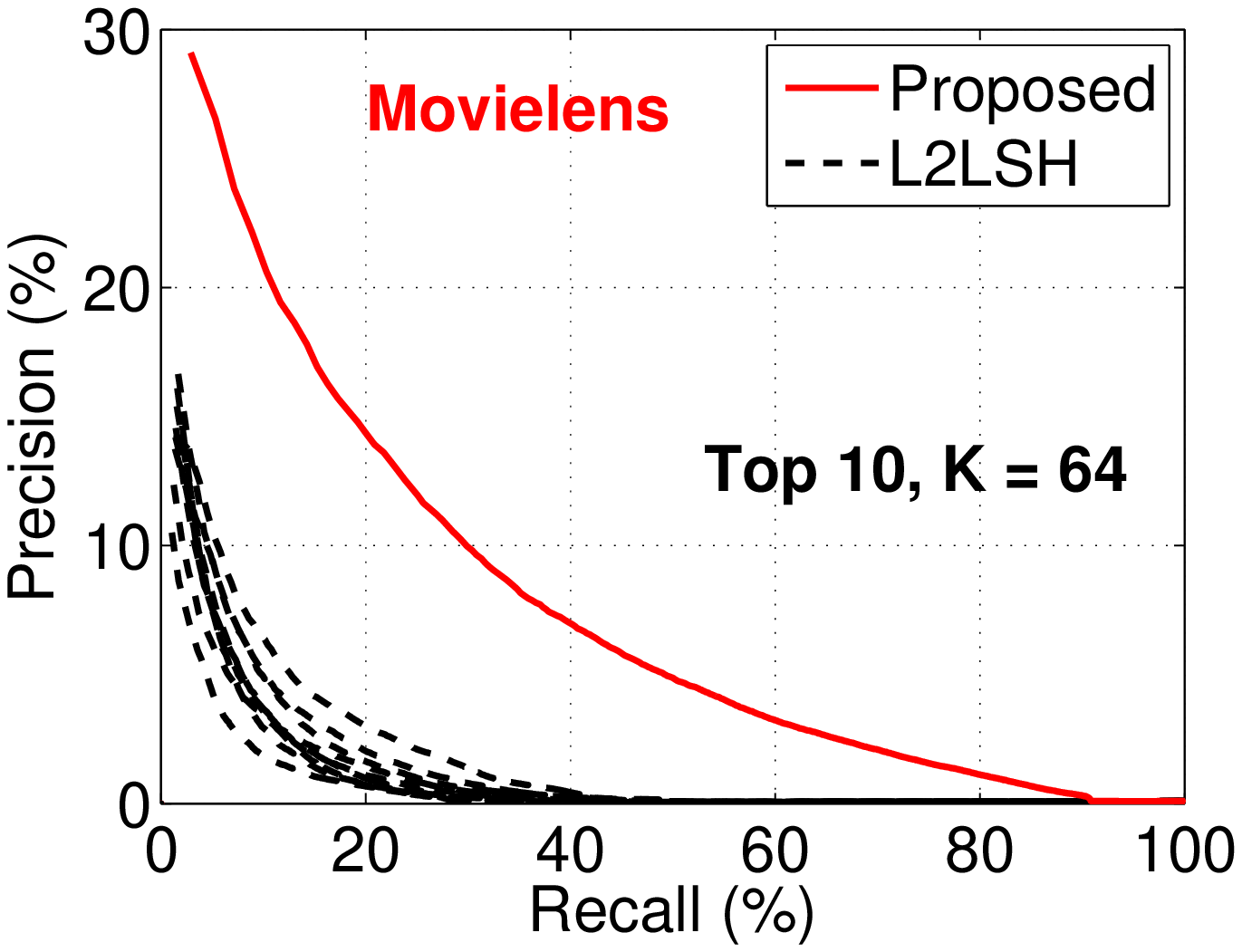}
}

\end{center}
\vspace{-0.2in}
\caption{\textbf{Movielens}. Precision-Recall curves (higher is better), of retrieving top-$T$ items, for $T=1, 5, 10$. We vary the number of hashes $K$ from 64 to 512. The proposed algorithm (solid, red if color is available) significantly outperforms L2LSH. We fix the parameters $m=3$, $U=0.83$, and $r=2.5$ for our proposed method and we present the results of L2LSH for all $r$ values in $\{1, 1.5, 2, 2.5, 3, 3.5, 4, 4.5, 5\}$. Because the difference between our method and L2LSH is large, we do not label  curves at different $r$ values for L2LSH.   }\label{fig_MovielensRanking}
\end{figure}

\begin{figure}[h!]
\begin{center}
\mbox{
\includegraphics[width=2.25in]{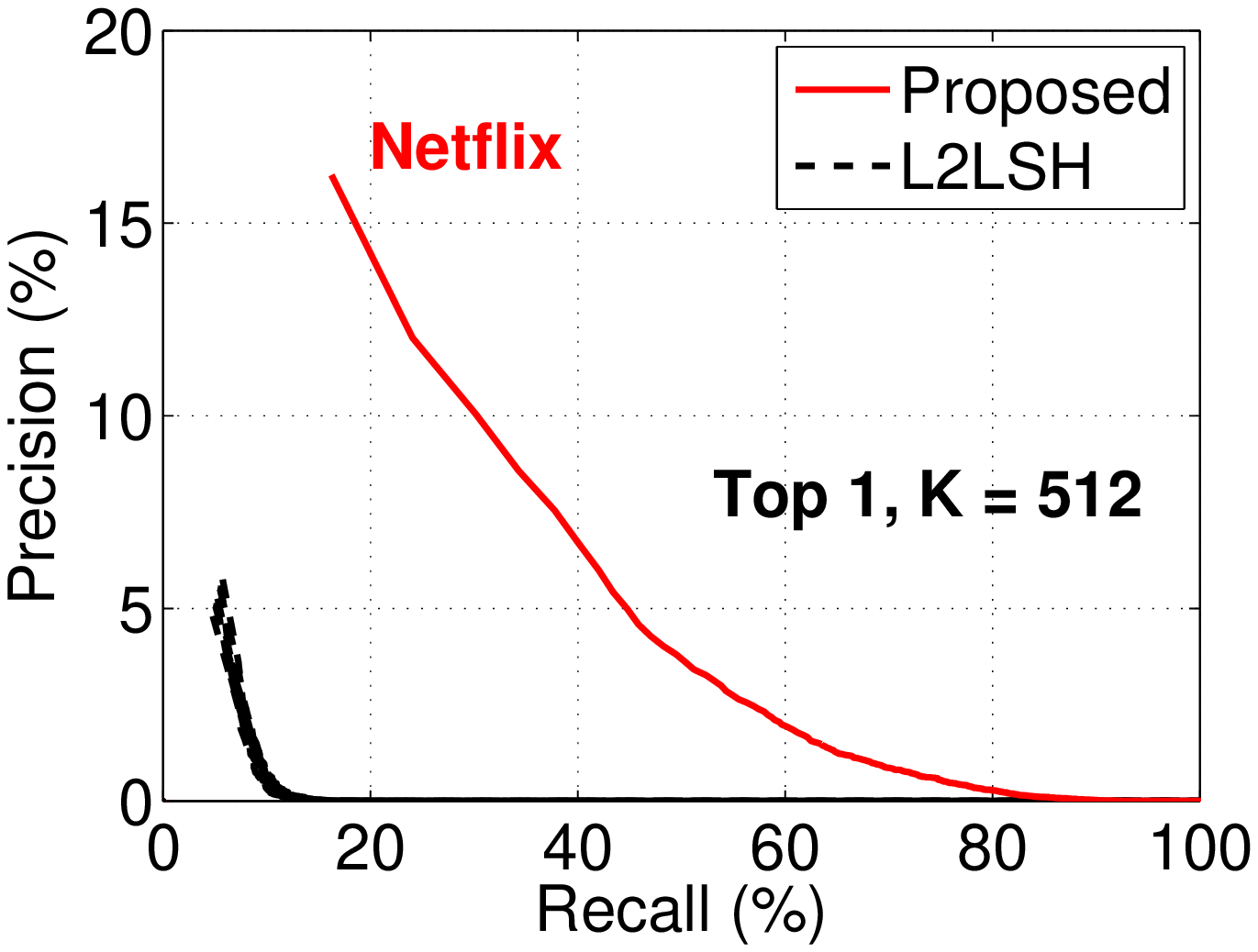}\hspace{-0.13in}
\includegraphics[width=2.25in]{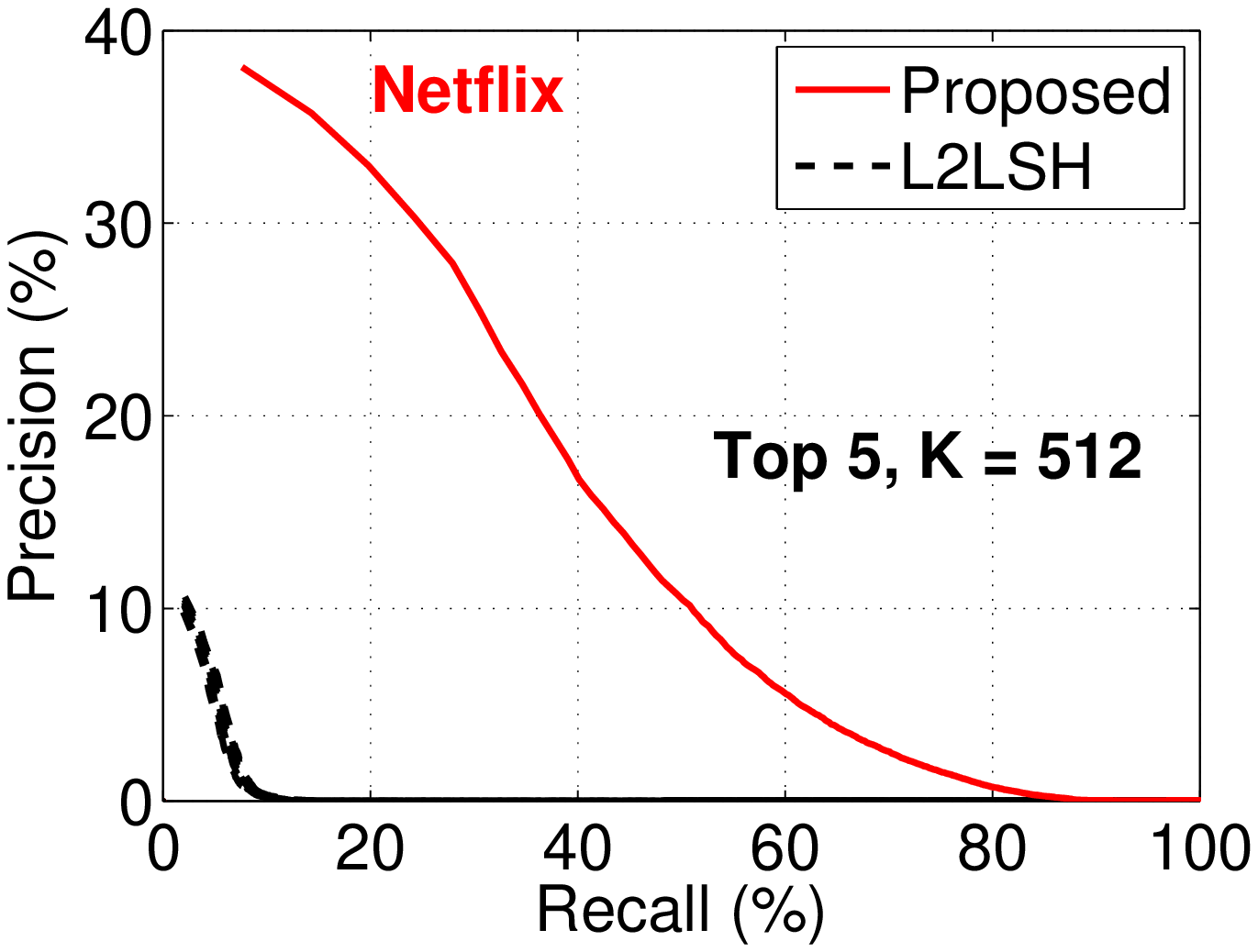}\hspace{-0.13in}
\includegraphics[width=2.25in]{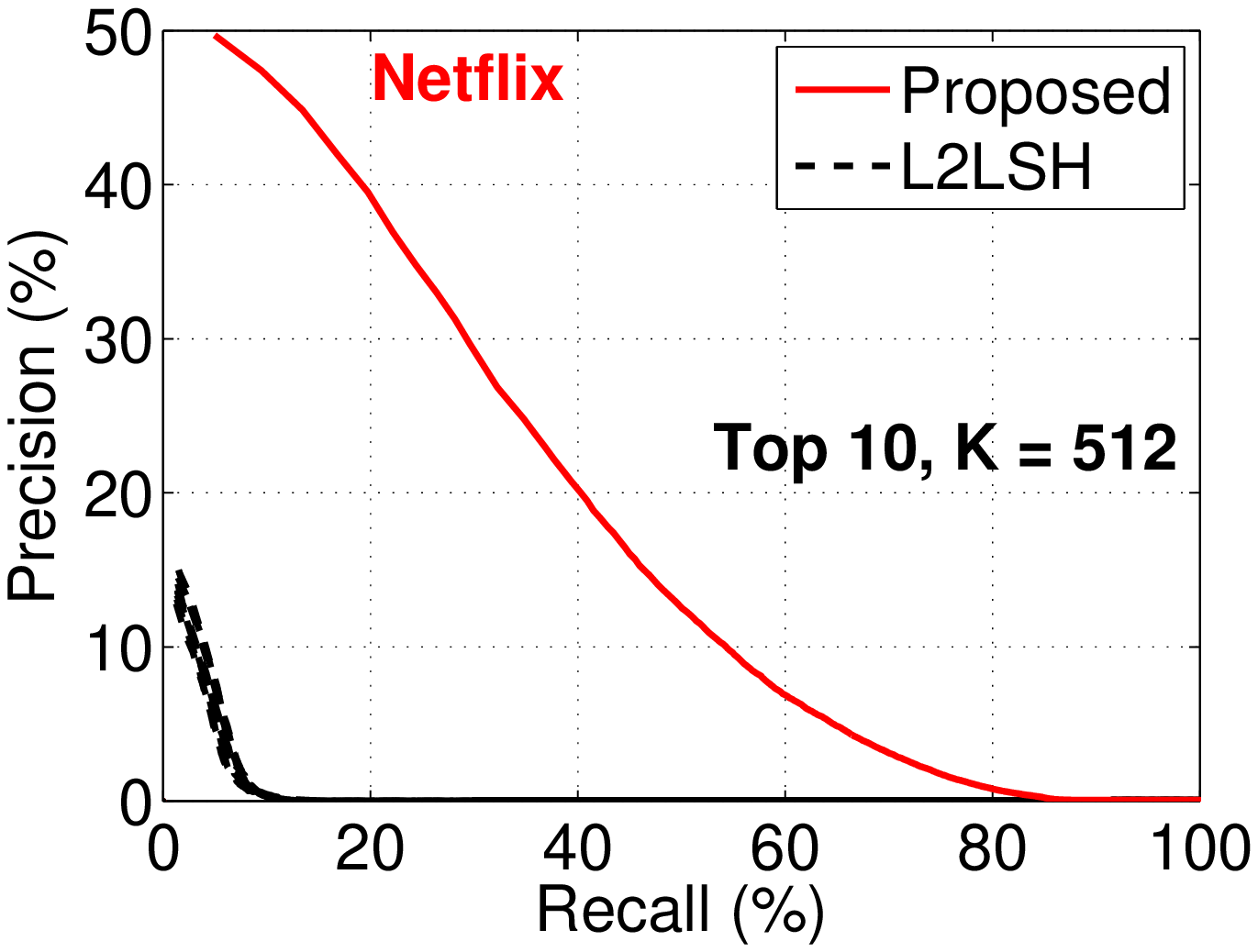}
}

\mbox{
\includegraphics[width=2.25in]{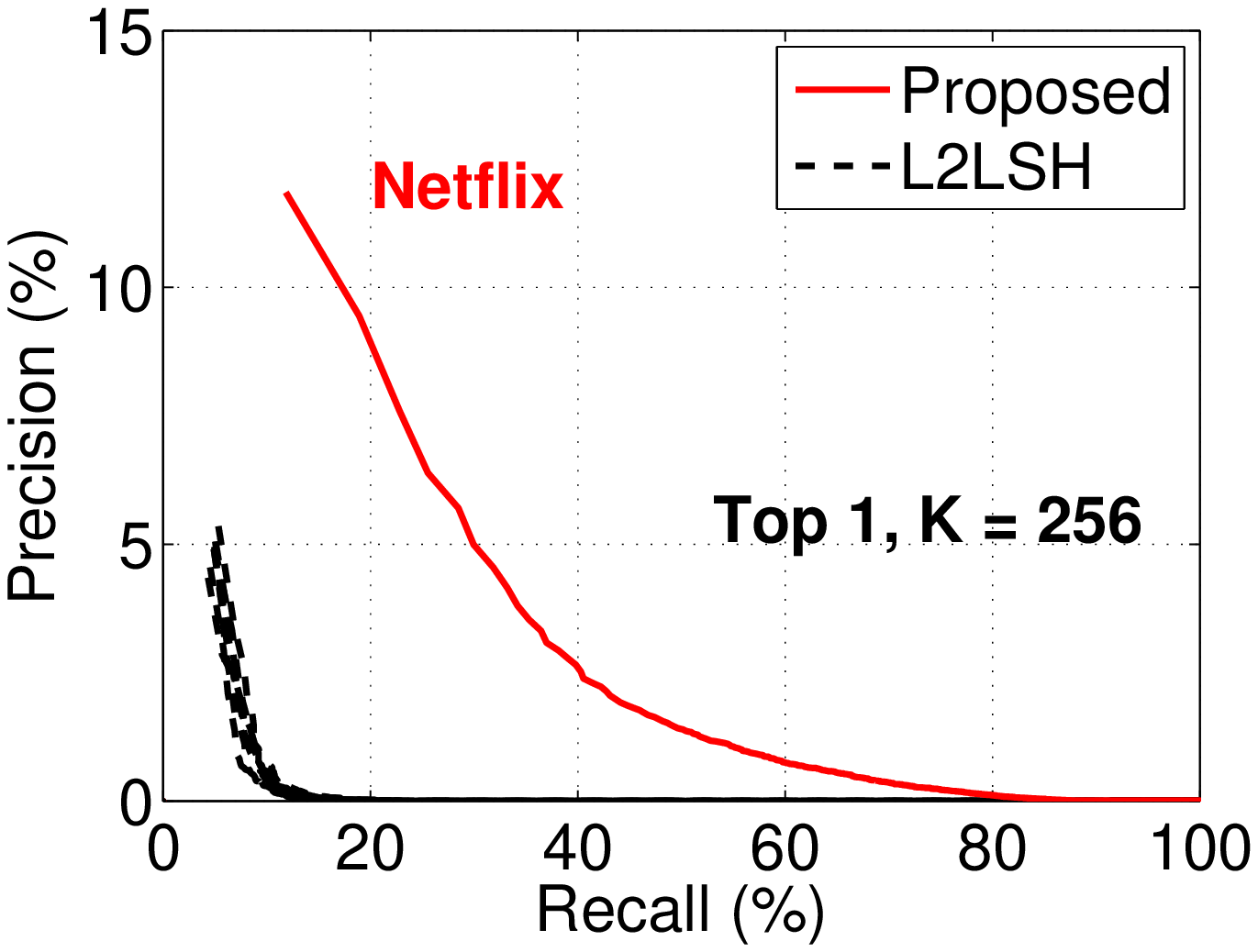}\hspace{-0.13in}
\includegraphics[width=2.25in]{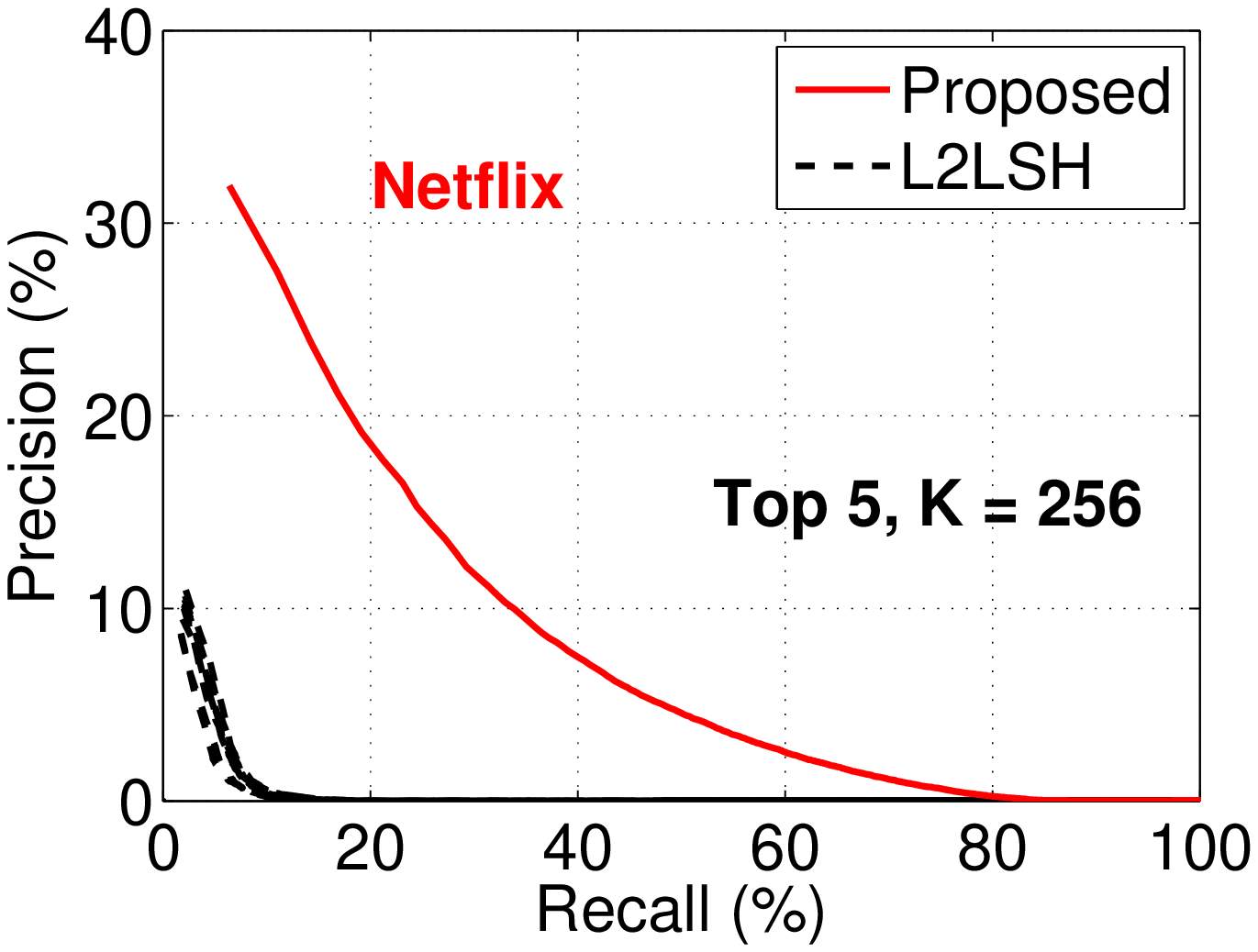}\hspace{-0.13in}
\includegraphics[width=2.25in]{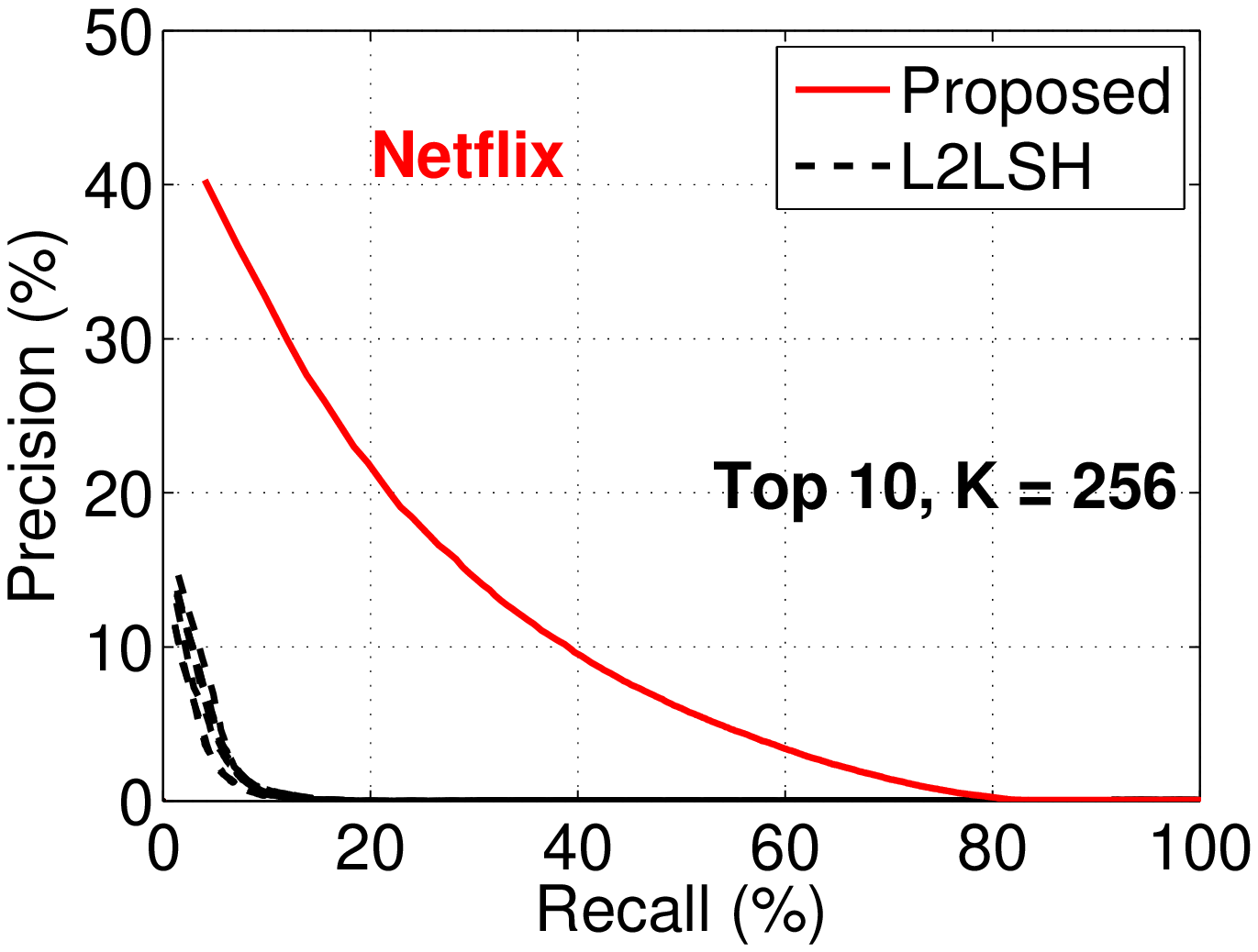}
}

\mbox{
\includegraphics[width=2.25in]{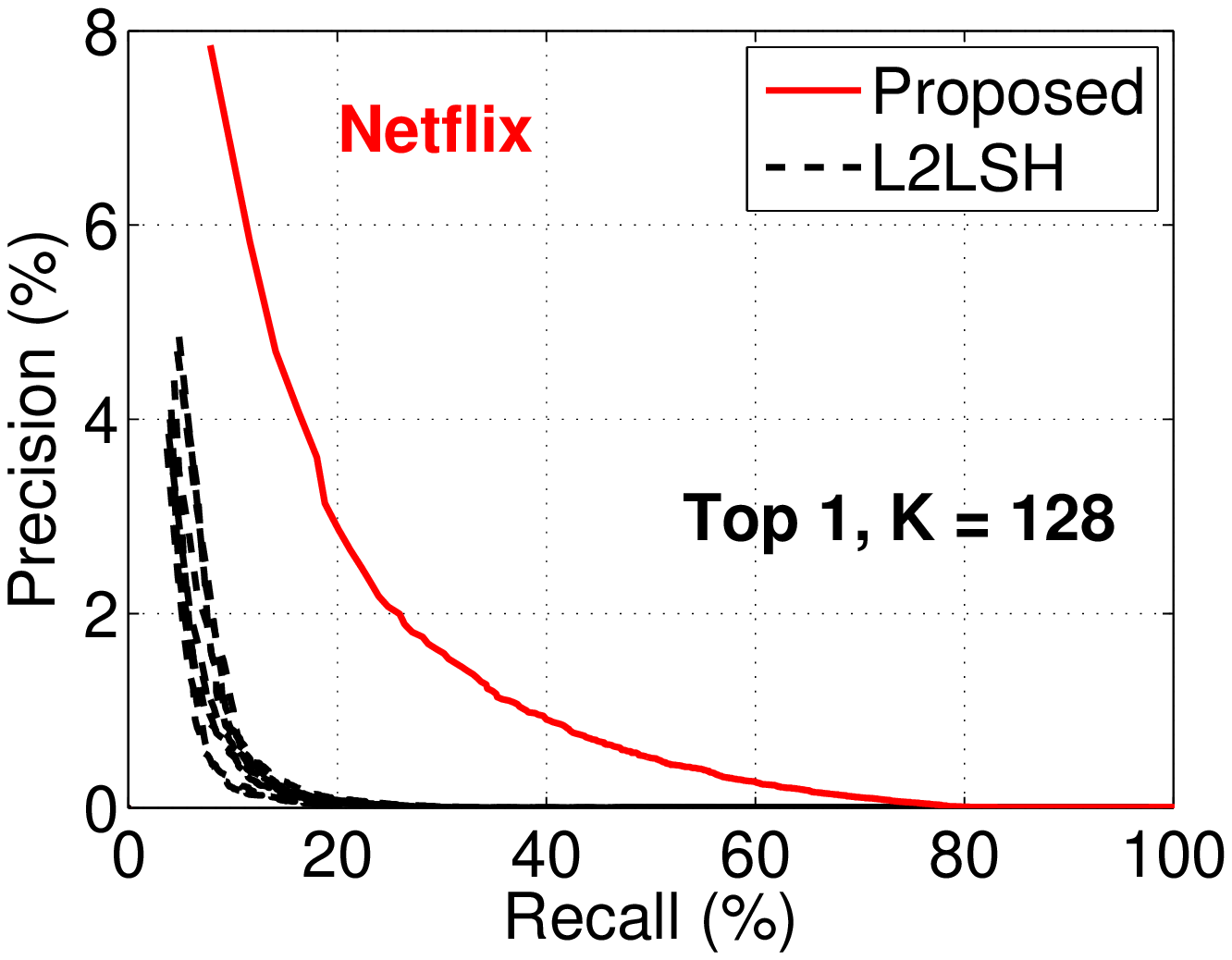}\hspace{-0.13in}
\includegraphics[width=2.25in]{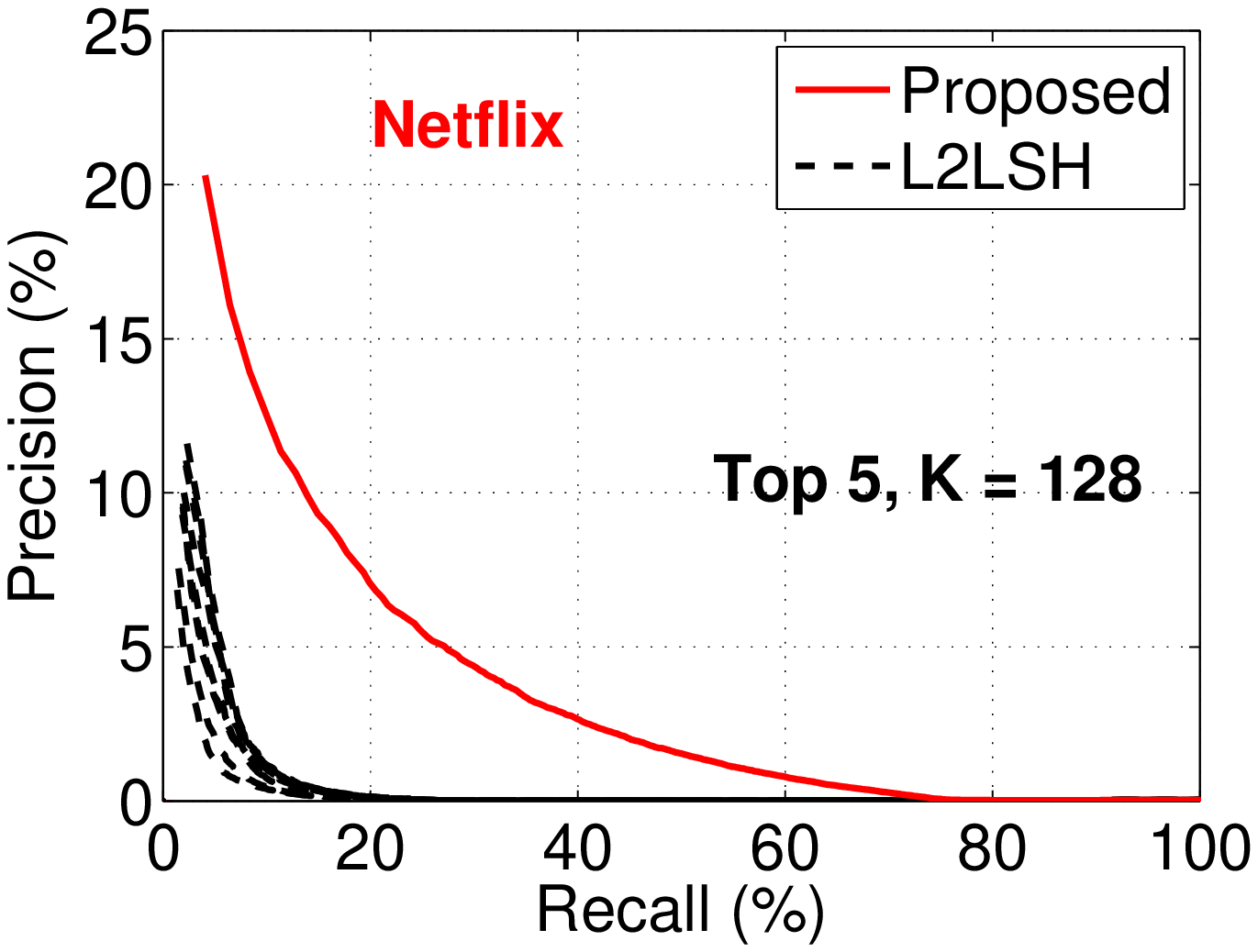}\hspace{-0.13in}
\includegraphics[width=2.25in]{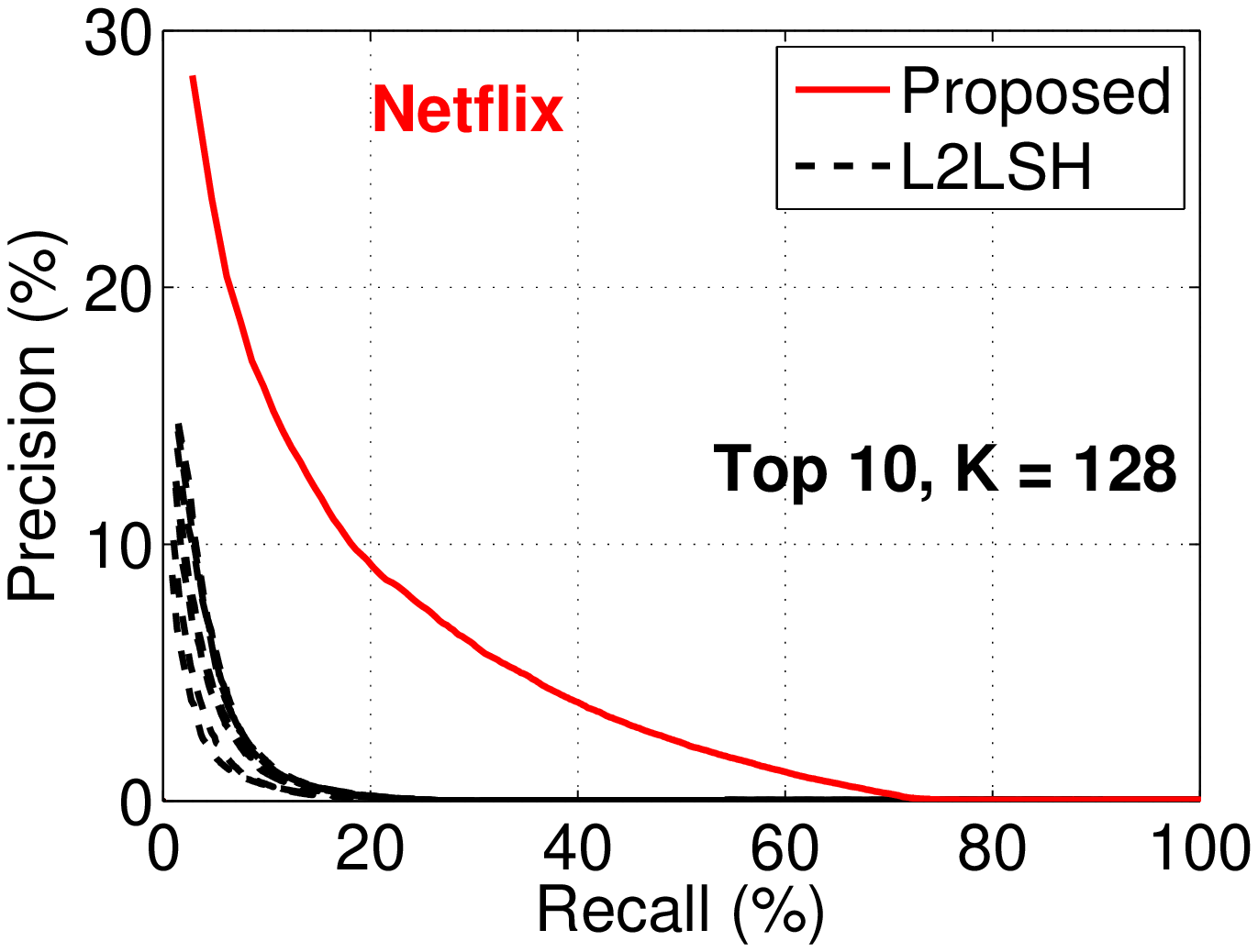}
}

\mbox{
\includegraphics[width=2.25in]{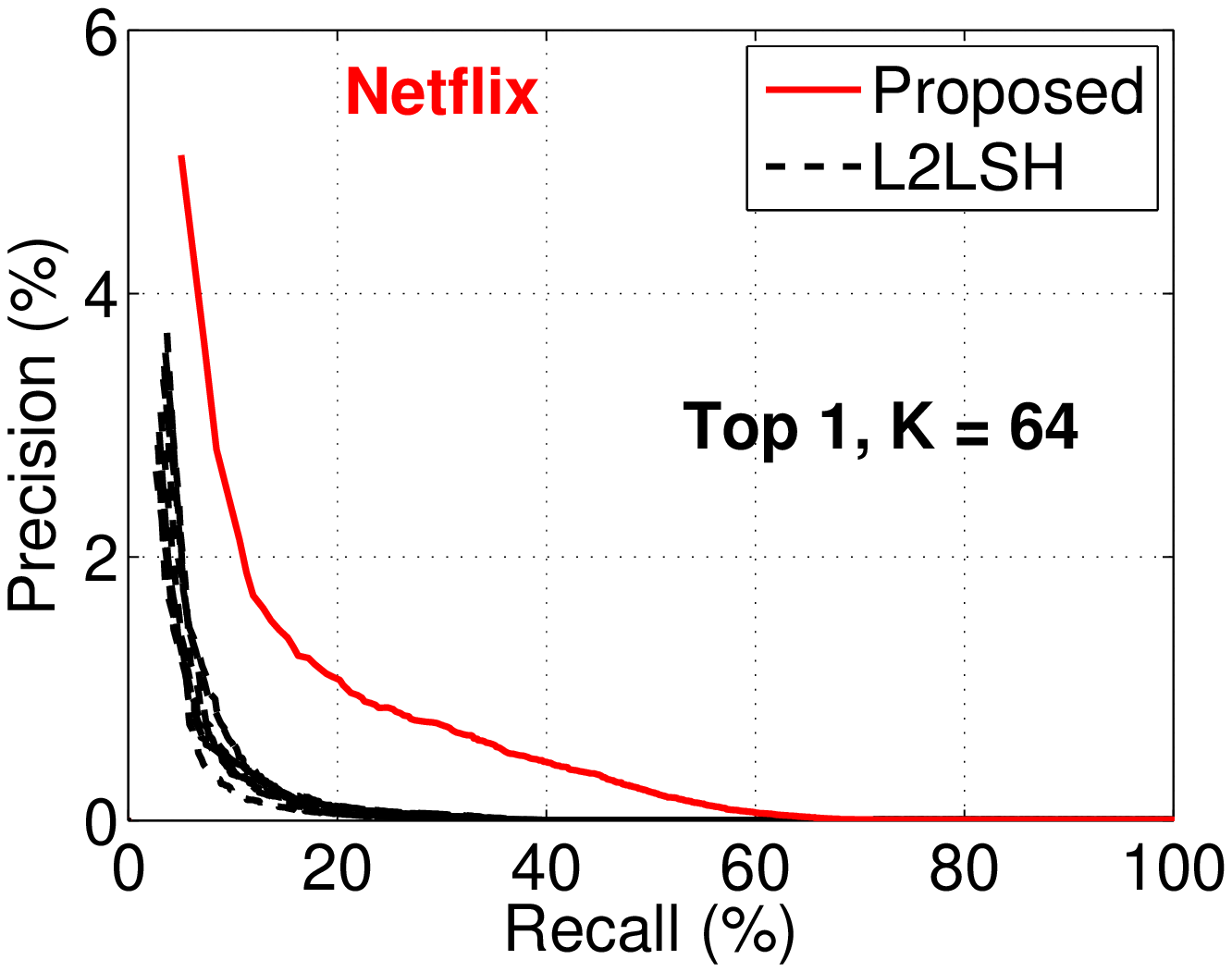}\hspace{-0.13in}
\includegraphics[width=2.25in]{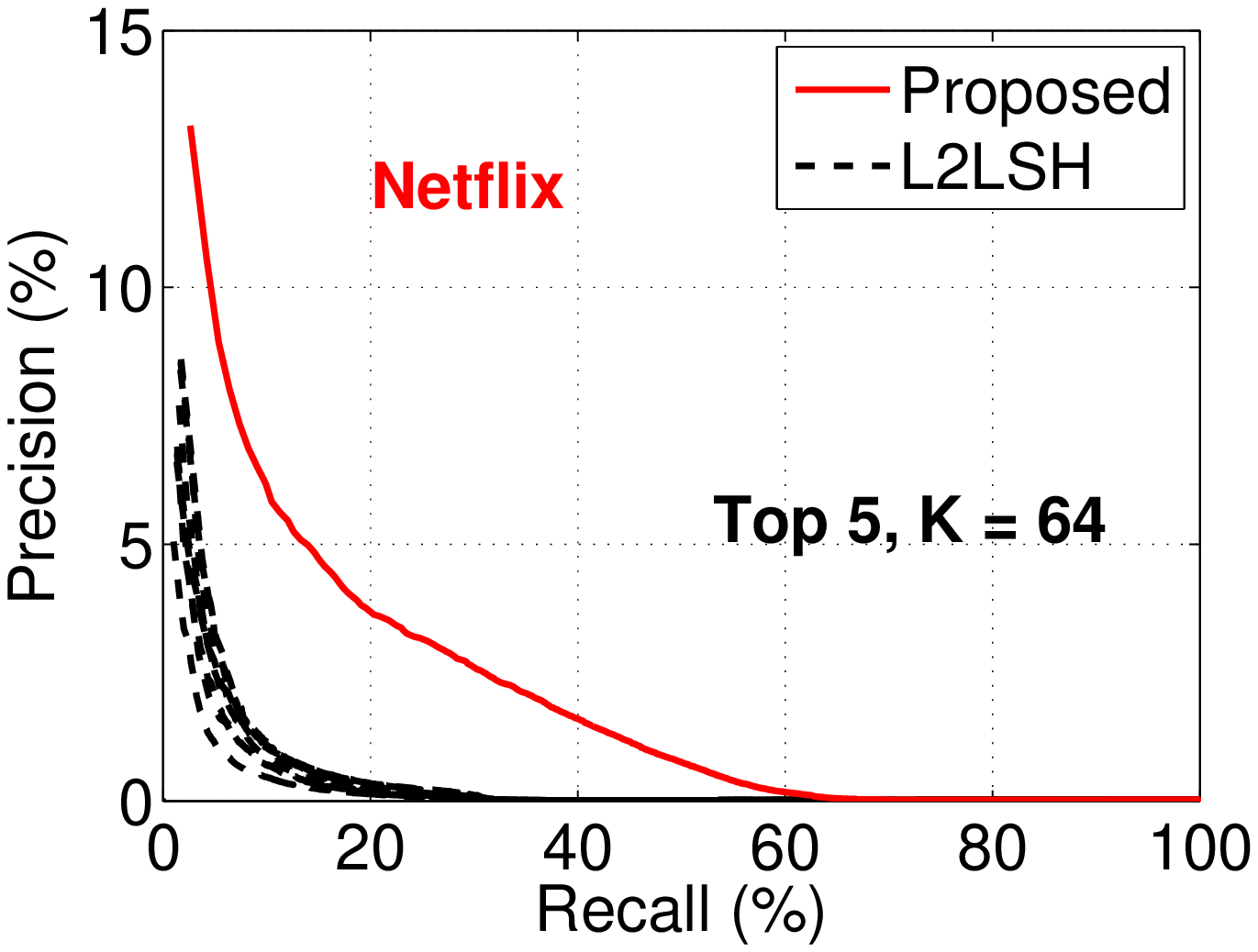}\hspace{-0.13in}
\includegraphics[width=2.25in]{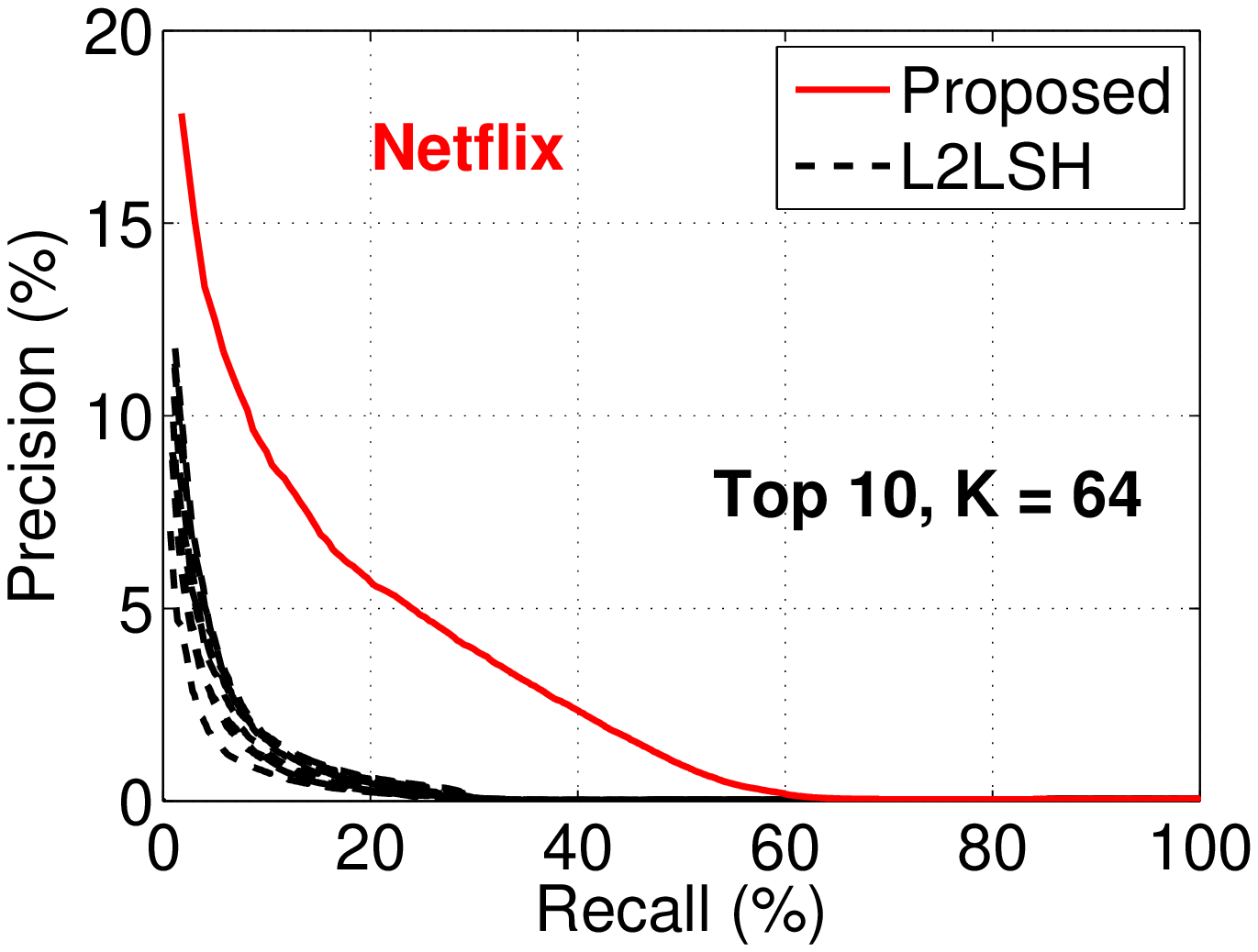}
}

\end{center}
\vspace{-0.2in}
\caption{\textbf{Netflix}. Precision-Recall curves (higher is better), of retrieving top-$T$ items, for $T=1, 5, 10$. We vary the number of hashes $K$ from 64 to 512. The proposed algorithm (solid, red if color is available) significantly outperforms L2LSH. We fix the parameters $m=3$, $U=0.83$, and $r=2.5$ for our proposed method and we present the results of L2LSH for all $r$ values in $\{1, 1.5, 2, 2.5, 3, 3.5, 4, 4.5, 5\}$. Because the difference between our method and L2LSH is large, we do not label  curves at different $r$ values for L2LSH.    }\label{fig_NetflixRanking}
\end{figure}

\noindent\textbf{Choice of $r$}. \ \ We need to specify this important parameter for our proposed algorithm as well as L2LSH. We have shown in Figure~\ref{fig:Rhom3} that, for our proposed algorithm, it is overall good to choose $m=3$, $U=0.83$ and $r=2.5$. This theoretical result  largely frees us from  the burden of choosing parameters. We still need to choose $r$ for L2LSH. To ensure that  L2LSH is not being suboptimal because of the choice of $r$, we report the results of L2LSH for all $r\in \{1, 1.5, 2, 2.5, 3, 3.5, 4, 4.5, 5\}$, and we show that our proposed method (with $r=2.5$) very significantly outperforms L2LSH at all choices of $r$, in Figure~\ref{fig_MovielensRanking} and Figure~\ref{fig_NetflixRanking}, for $K=64, 128, 256, 512$ hashes.

The good performance of our algorithm shown in Figure~\ref{fig_MovielensRanking} and Figure~\ref{fig_NetflixRanking} is not surprising because we know from Theorem~\ref{theo:collprobnew} that the collision under the new hash function is a direct indicator of high inner product. As the number of hash codes is increased, the performance of the proposed methodology shows bigger improvements over L2LSH.  The suboptimal performance of L2LSH clearly indicates that the norms of the item characteristic vectors do play a significant role in item recommendation task. Also, the experiments clearly establishes the need of proposed asymmetric transformation $P$ and $Q$.

\newpage\clearpage

\begin{figure}[h!]
\begin{center}
\mbox{
\includegraphics[width=2.25in]{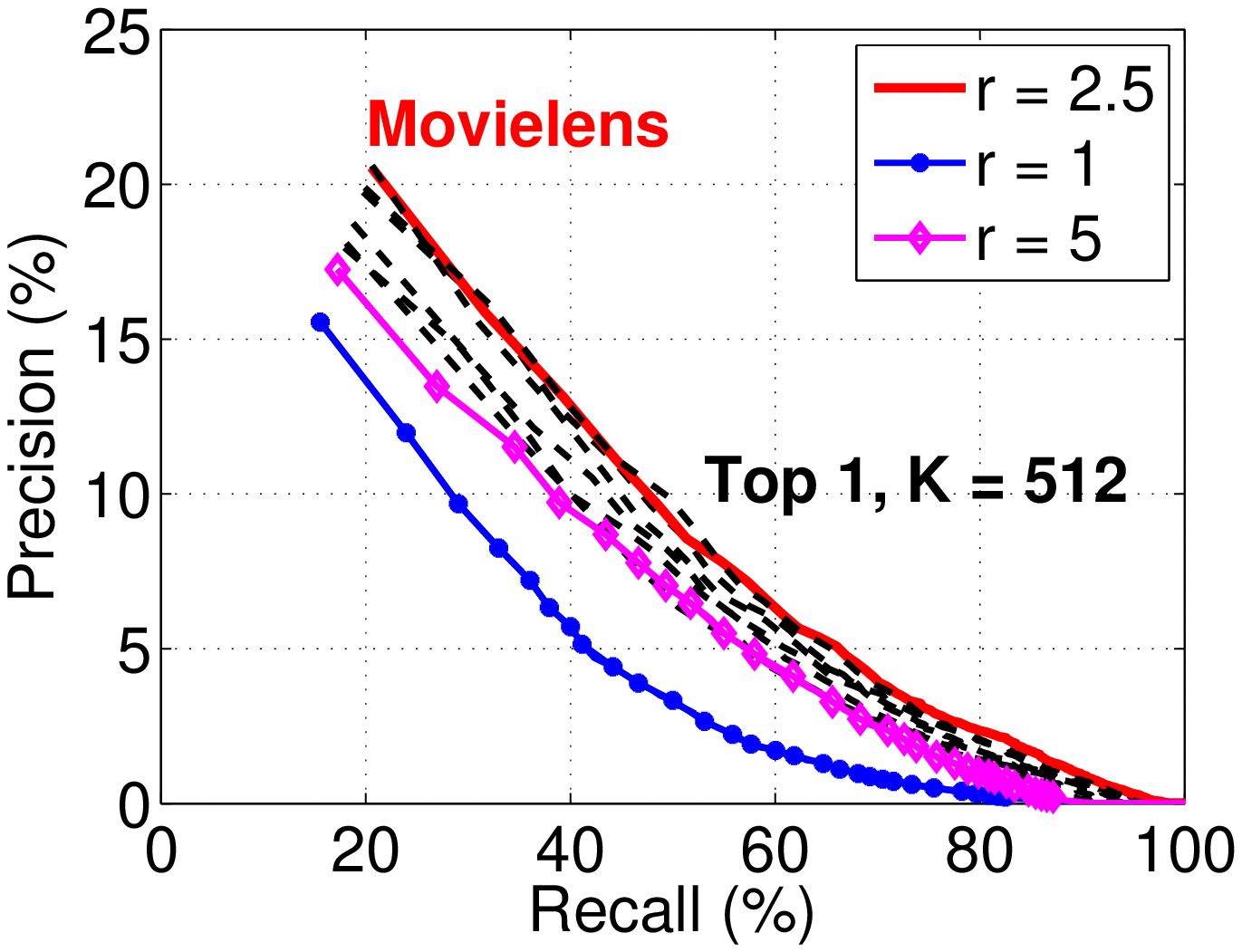}\hspace{-0.13in}
\includegraphics[width=2.25in]{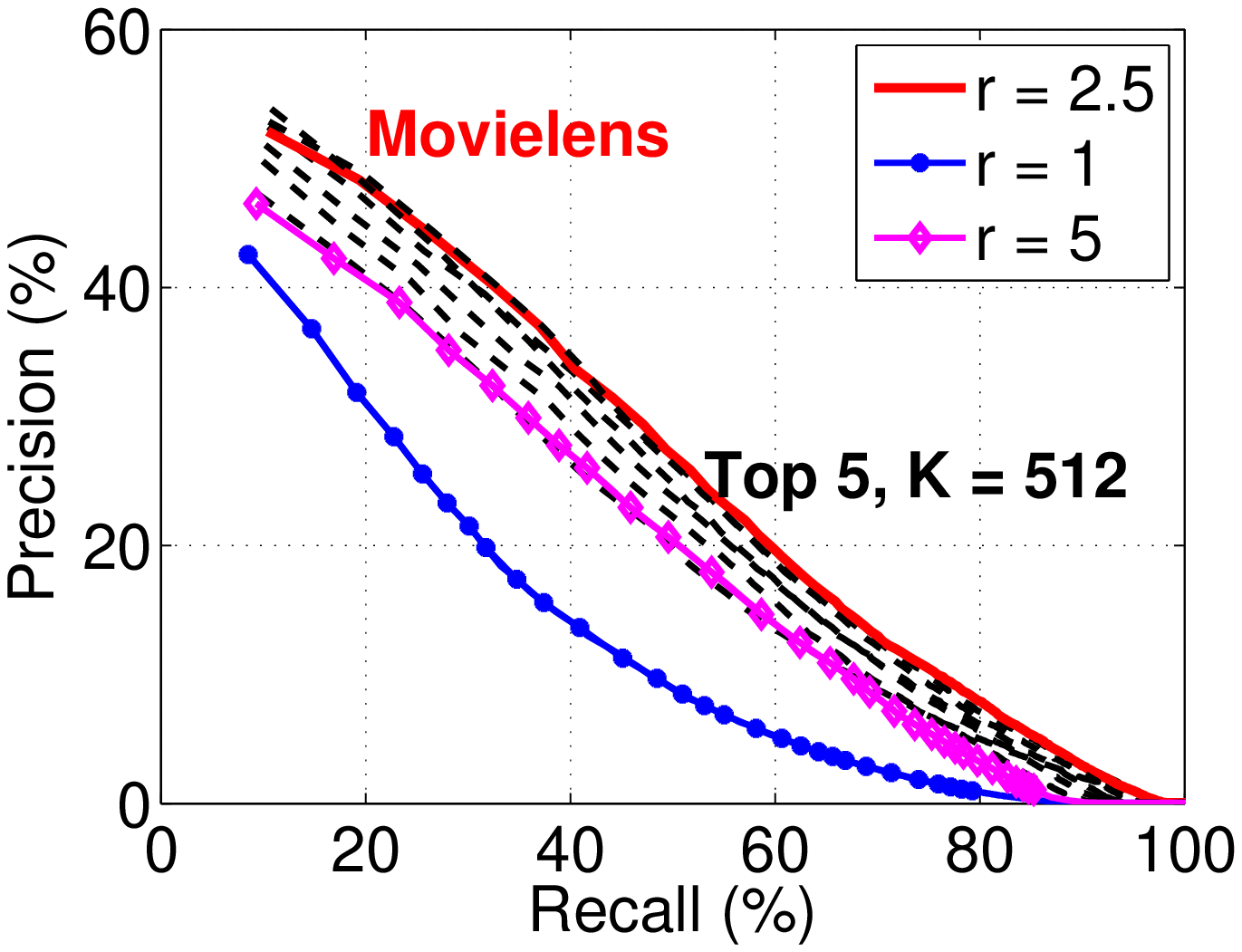}\hspace{-0.13in}
\includegraphics[width=2.25in]{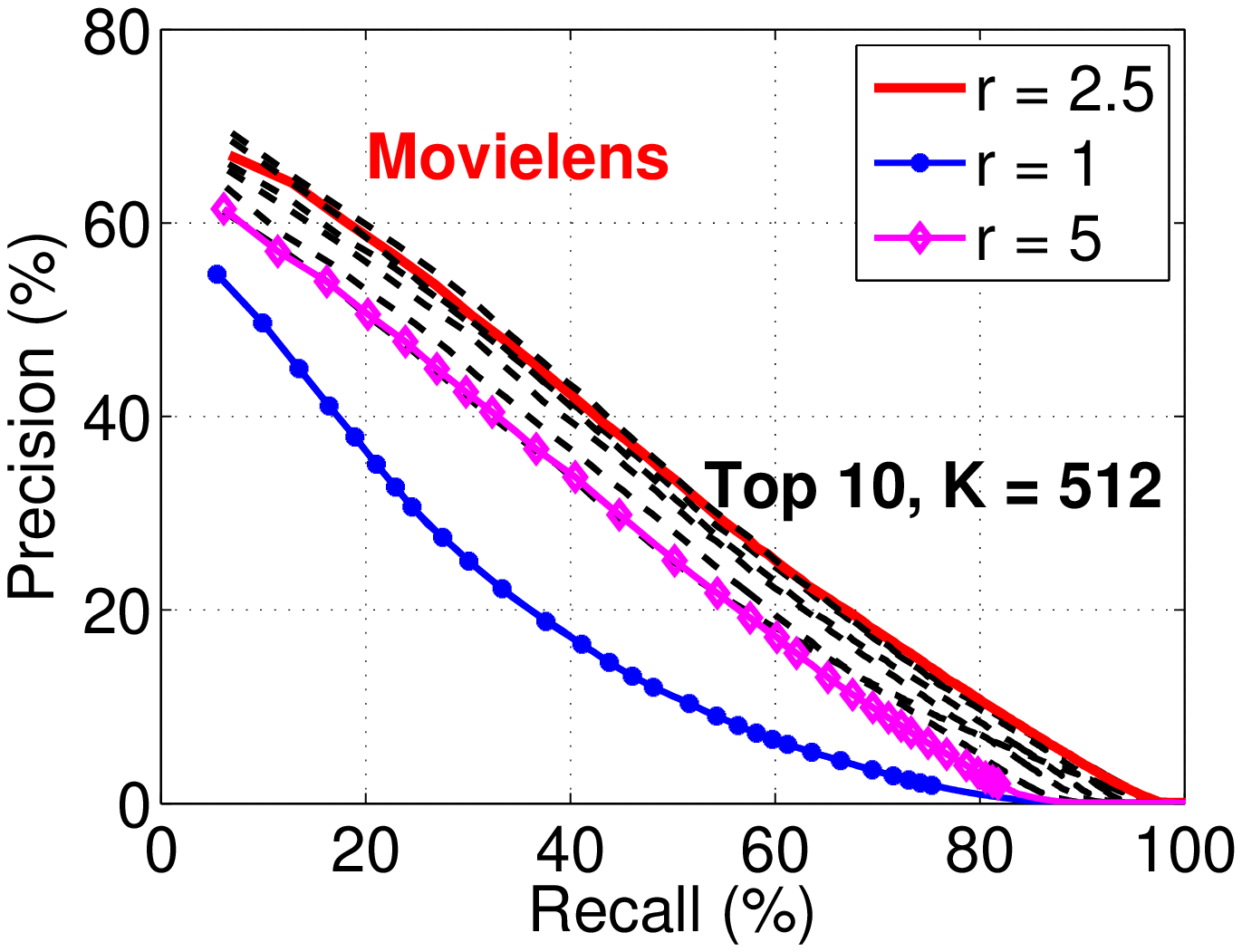}
}
\mbox{
\includegraphics[width=2.25in]{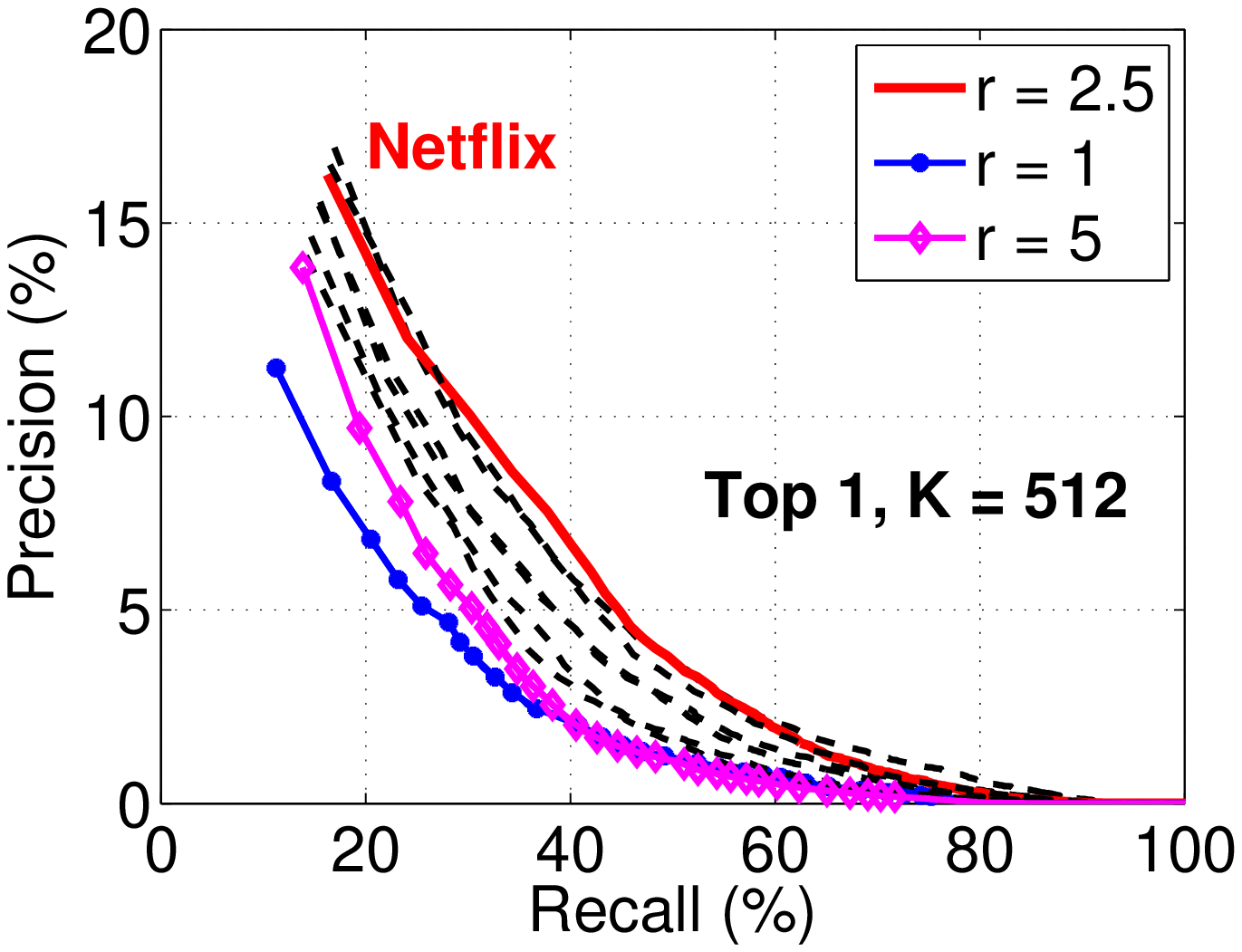}\hspace{-0.13in}
\includegraphics[width=2.25in]{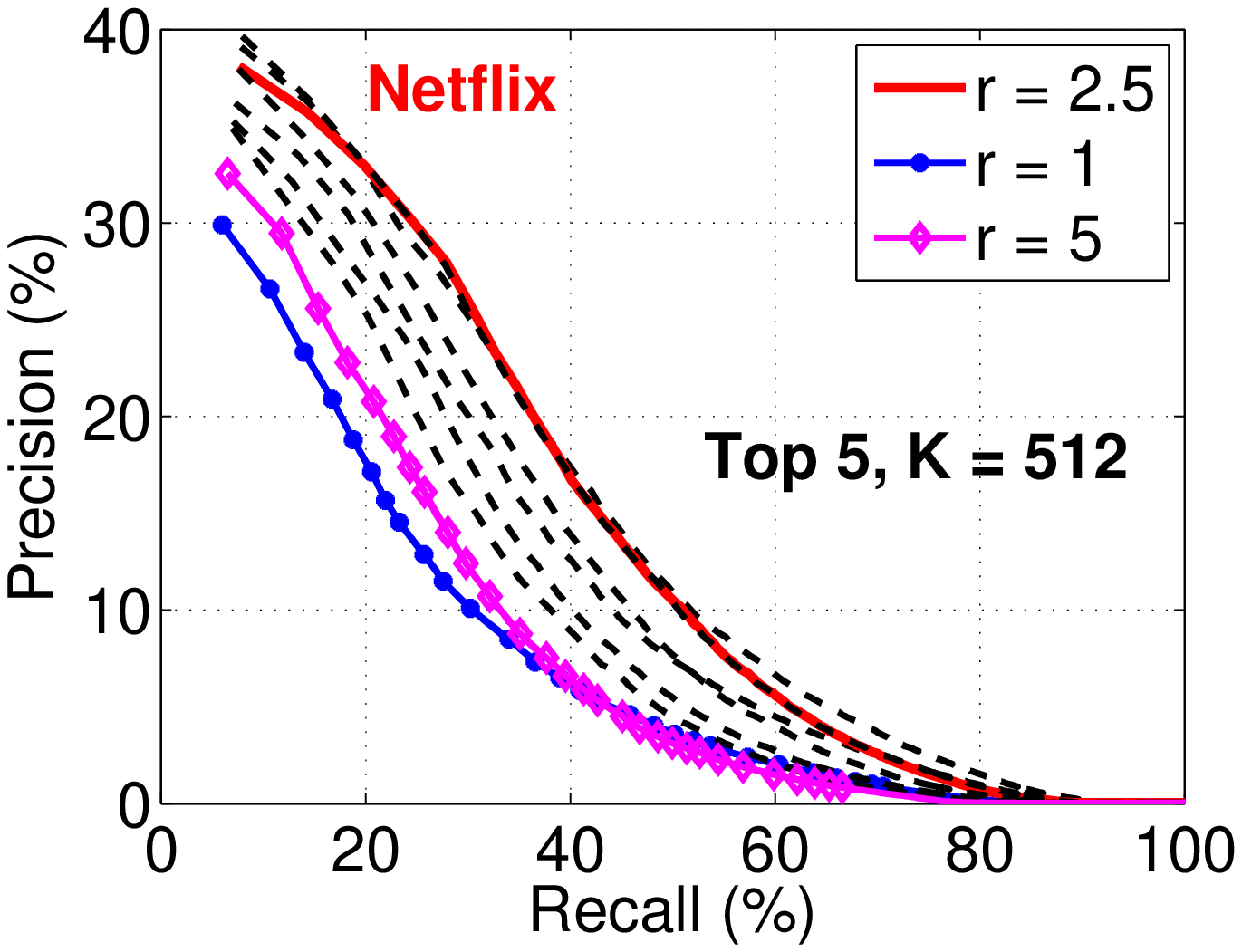}\hspace{-0.13in}
\includegraphics[width=2.25in]{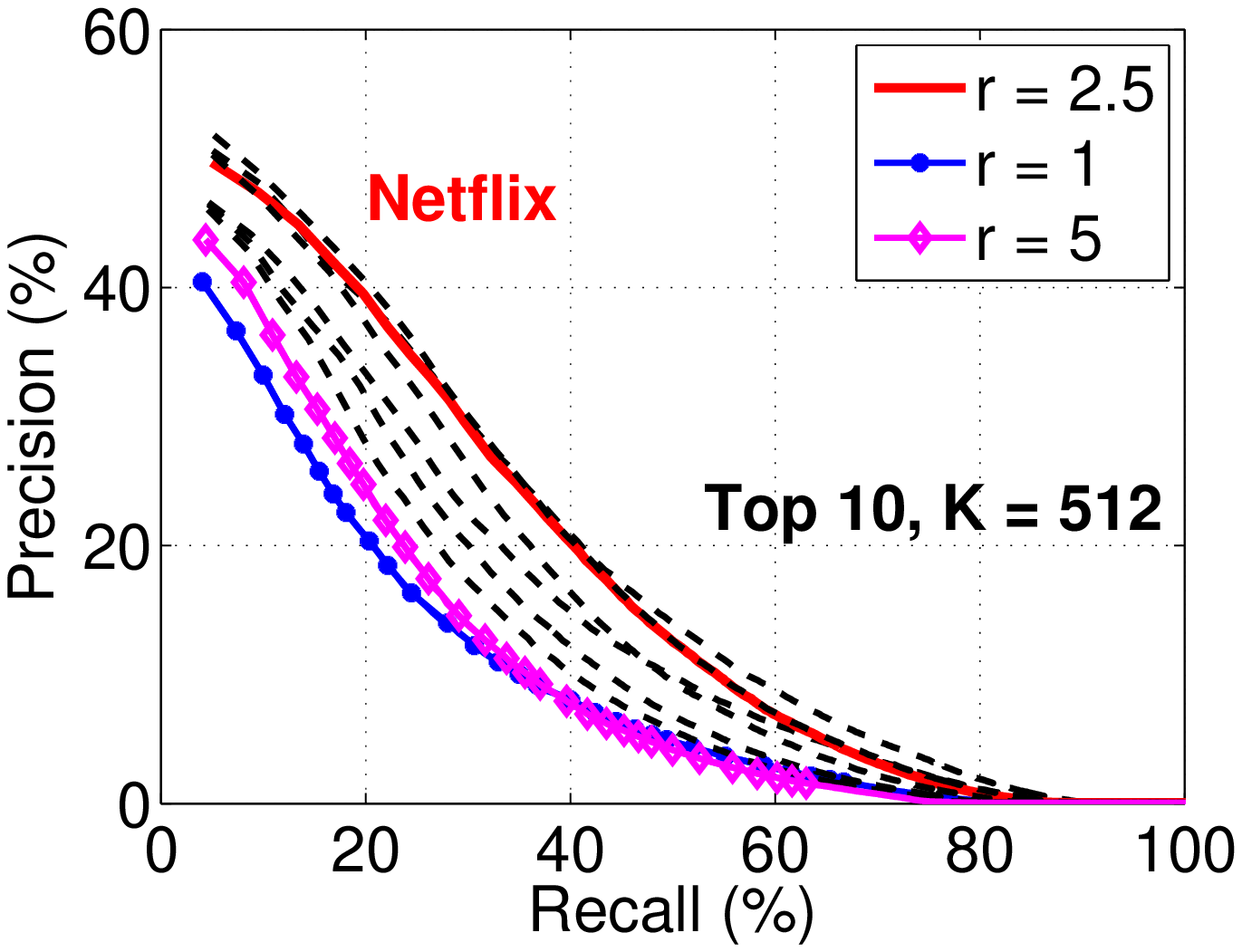}
}

\end{center}
\vspace{-0.2in}
\caption{Impact of  $r$ on the performance of the proposed method. We present the precision-recall curves  for $m=3$, $U=0.83$, and $r \in \{1, 1.5, 2, 2.5, 3, 3.5, 4, 4.5, 5\}$, which   demonstrate that, (i) $r=2.5$ is indeed a good choice; and (ii) the performance is not too sensitive to $r$ unless it is much away from 2.5.  }\label{fig_RankingProp}
\end{figure}

To close this section, we present Figure~\ref{fig_RankingProp} to visualize the impact of the parameter $r$ on the performance of our proposed method. Our theory and Figure~\ref{fig:Rhom3}  have already shown that we can achieve close to optimal performance by choosing $m=3$, $U=0.83$, and $r=2.5$. Nevertheless, it is still interesting to see how the theory is confirmed by experiments. Figure~\ref{fig_RankingProp} presents the precision-recall curves for our proposed method with $m=3$, $U=0.83$, and $r \in \{1, 1.5, 2, 2.5, 3, 3.5, 4, 4.5, 5\}$. For clarify, we only differentiate $r = 1, 2.5, 5$ from the rest of the curves. The results demonstrate that $r=2.5$ is indeed a good choice and the performance is not too sensitive to $r$ unless it is much away from 2.5.

\section{Conclusion and Future Work}

MIPS (maximum inner product search) naturally arises in numerous practical scenarios, e.g., collaborative filtering. Given a query data vector, the task of MIPS is to find data vectors from the repository which are most similar to the query in terms of (un-normalized) inner product (instead of distance).  This problem is challenging and, prior to our work, there existed no provably sublinear time algorithms for MIPS. The current framework of LSH (locality sensitive hashing) is not sufficient for solving MIPS. \\

In this study,  we develop {\em ALSH} (asymmetric LSH), which generalizes the existing LSH framework by applying (appropriately chosen) asymmetric transformations to the input query  vector and the data vectors in the repository. We present an implementation of ALSH by proposing a novel transformation which converts the original inner products into L2 distances in the transformed space.  We demonstrate, both theoretically and empirically, that this implementation of ALSH provides a provably efficient solution to MIPS. \\

\newpage

We believe our work will lead to several interesting (and practically useful) lines of \textbf{future work}:
\begin{itemize}

\item {\em Three-way (and higher-order) maximum inner product search:}\hspace{0.1in} In this paper, we have focused on pairwise similarity search. Three-way (or even higher-order) search can also be important in practice and might  attract more attentions in the near future due to the recent pilot studies on efficient three-way similarity computation~\cite{Article:Li_Church_CL07,Proc:Li_Konig_NIPS10,Proc:Shrivastava_NIPS13}. Extending ALSH to  three-way MIPS could be very useful.

\item {\em Other efficient similarities:}\hspace{0.1in}  Finding other similarities for which there exist fast retrieval algorithms is an important problem~\cite{Proc:Chierichetti_SODA12}. Exploring other similarity functions, which can be efficient solved using  asymmetric hashing  is an interesting future area. One good example is to find special ALSH schemes for binary data by exploring prior powerful hashing methods for binary data such as {\em (b-bit) minwise hashing} and {\em one permutation hashing}~\cite{Proc:Broder,Proc:Li_Owen_Zhang_NIPS12}.

\item {\em  Fast hashing for MIPS:}\hspace{0.1in} Our proposed hash function uses random projection as the main hashing scheme. There is a rich set of literature~\cite{Proc:Li_Hastie_Church_KDD06,Proc:Ailon_STOC06,Proc:Dasgupta_KDD11,Proc:Li_Owen_Zhang_NIPS12,Proc:OneHashLSH_ICML14} on making hashing  faster. Evaluations of these new faster techniques will further improve the runtime guarantees in this paper.

\item {\em Applications:}\hspace{0.1in} We have  evaluated the efficiency of our scheme in the collaborative filtering application. An interesting set of future work consist of applying this hashing scheme for other applications mentioned in Section~\ref{sec:intro}. In particular, it will be exciting to apply efficient MIPS subroutines to improve the DPM based object detection~\cite{Proc:Dean_CVPR2013} and structural SVMs~\cite{Article:joachims_2009}.
\end{itemize}

{

}


\begin{thebibliography}{10}

\bibitem{Proc:Ailon_STOC06}
N.~Ailon and B.~Chazelle.
\newblock Approximate nearest neighbors and the fast
  \text{Johnson-Lindenstrauss} transform.
\newblock In {\em STOC}, pages 557--563, Seattle, WA, 2006.

\bibitem{Report:E2LSH}
A.~Andoni and P.~Indyk.
\newblock E2lsh: Exact euclidean locality sensitive hashing.
\newblock Technical report, 2004.

\bibitem{Proc:Broder}
A.~Z. Broder.
\newblock On the resemblance and containment of documents.
\newblock In {\em the Compression and Complexity of Sequences}, pages 21--29,
  Positano, Italy, 1997.

\bibitem{Proc:Charikar}
M.~S. Charikar.
\newblock Similarity estimation techniques from rounding algorithms.
\newblock In {\em STOC}, pages 380--388, Montreal, Quebec, Canada, 2002.

\bibitem{Proc:Chierichetti_SODA12}
F.~Chierichetti and R.~Kumar.
\newblock Lsh-preserving functions and their applications.
\newblock In {\em SODA}, pages 1078--1094, 2012.

\bibitem{Proc:Cremonesi_RecSys}
P.~Cremonesi, Y.~Koren, and R.~Turrin.
\newblock Performance of recommender algorithms on top-n recommendation tasks.
\newblock In {\em Proceedings of the fourth ACM conference on Recommender
  systems}, pages 39--46. ACM, 2010.

\bibitem{Proc:Das_WWW07}
A.~S. Das, M.~Datar, A.~Garg, and S.~Rajaram.
\newblock Google news personalization: scalable online collaborative filtering.
\newblock In {\em Proceedings of the 16th international conference on World
  Wide Web}, pages 271--280. ACM, 2007.

\bibitem{Proc:Dasgupta_KDD11}
A.~Dasgupta, R.~Kumar, and T.~Sarl{\'o}s.
\newblock Fast locality-sensitive hashing.
\newblock In {\em KDD}, pages 1073--1081, 2011.

\bibitem{Proc:Datar_SCG04}
M.~Datar, N.~Immorlica, P.~Indyk, and V.~S. Mirrokn.
\newblock Locality-sensitive hashing scheme based on $p$-stable distributions.
\newblock In {\em SCG}, pages 253 -- 262, Brooklyn, NY, 2004.

\bibitem{Proc:Dean_CVPR2013}
T.~Dean, M.~A. Ruzon, M.~Segal, J.~Shlens, S.~Vijayanarasimhan, and J.~Yagnik.
\newblock Fast, accurate detection of 100,000 object classes on a single
  machine.
\newblock In {\em Computer Vision and Pattern Recognition (CVPR), 2013 IEEE
  Conference on}, pages 1814--1821. IEEE, 2013.

\bibitem{Article:Felzenszwalb_PAMI2010}
P.~F. Felzenszwalb, R.~B. Girshick, D.~McAllester, and D.~Ramanan.
\newblock Object detection with discriminatively trained part-based models.
\newblock {\em Pattern Analysis and Machine Intelligence, IEEE Transactions
  on}, 32(9):1627--1645, 2010.

\bibitem{Article:Friedman_74}
J.~H. Friedman and J.~W. Tukey.
\newblock A projection pursuit algorithm for exploratory data analysis.
\newblock {\em IEEE Transactions on Computers}, 23(9):881--890, 1974.

\bibitem{Article:LSH_12}
S.~Har-Peled, P.~Indyk, and R.~Motwani.
\newblock Approximate nearest neighbor: Towards removing the curse of
  dimensionality.
\newblock {\em Theory of Computing}, 8(14):321--350, 2012.

\bibitem{Proc:Henzinger_SIGIR06}
M.~R. Henzinger.
\newblock Finding near-duplicate web pages: a large-scale evaluation of
  algorithms.
\newblock In {\em SIGIR}, pages 284--291, 2006.

\bibitem{Proc:Indyk_STOC98}
P.~Indyk and R.~Motwani.
\newblock Approximate nearest neighbors: Towards removing the curse of
  dimensionality.
\newblock In {\em STOC}, pages 604--613, Dallas, TX, 1998.

\bibitem{Article:joachims_2009}
T.~Joachims, T.~Finley, and C.-N.~J. Yu.
\newblock Cutting-plane training of structural svms.
\newblock {\em Machine Learning}, 77(1):27--59, 2009.

\bibitem{Proc:Koenigstein_CIKM12}
N.~Koenigstein, P.~Ram, and Y.~Shavitt.
\newblock Efficient retrieval of recommendations in a matrix factorization
  framework.
\newblock In {\em CIKM}, pages 535--544, 2012.

\bibitem{Proc:Koren_KDD2008}
Y.~Koren.
\newblock Factorization meets the neighborhood: a multifaceted collaborative
  filtering model.
\newblock In {\em KDD}, pages 426--434. ACM, 2008.

\bibitem{Article:Koren_2009}
Y.~Koren, R.~Bell, and C.~Volinsky.
\newblock Matrix factorization techniques for recommender systems.

\bibitem{Article:Li_Church_CL07}
P.~Li and K.~W. Church.
\newblock A sketch algorithm for estimating two-way and multi-way associations.
\newblock {\em Computational Linguistics (Preliminary results appeared in
  HLT/EMNLP 2005)}, 33(3):305--354, 2007.

\bibitem{Proc:Li_Hastie_Church_KDD06}
P.~Li, T.~J. Hastie, and K.~W. Church.
\newblock Very sparse random projections.
\newblock In {\em KDD}, pages 287--296, Philadelphia, PA, 2006.

\bibitem{Proc:Li_Konig_NIPS10}
P.~Li, A.~C. {K\"{o}nig}, and W.~Gui.
\newblock b-bit minwise hashing for estimating three-way similarities.
\newblock In {\em Advances in Neural Information Processing Systems},
  Vancouver, BC, 2010.

\bibitem{Report:RPCodeLSH2014}
P.~Li, M.~Mitzenmacher, and A.~Shrivastava.
\newblock Coding for random projections and approximate near neighbor search.
\newblock Technical report, arXiv:1403.8144, 2014.

\bibitem{Proc:Li_Owen_Zhang_NIPS12}
P.~Li, A.~B. Owen, and C.-H. Zhang.
\newblock One permutation hashing.
\newblock In {\em NIPS}, Lake Tahoe, NV, 2012.

\bibitem{Proc:Manku_WWW07}
G.~S. Manku, A.~Jain, and A.~D. Sarma.
\newblock {D}etecting {N}ear-{D}uplicates for {W}eb-{C}rawling.
\newblock In {\em WWW}, Banff, Alberta, Canada, 2007.

\bibitem{Book:Raj_Ullman}
A.~Rajaraman and J.~Ullman.
\newblock {\em Mining of Massive Datasets}.
\newblock http://i.stanford.edu/~ullman/mmds.html.

\bibitem{Proc:Ram_KDD12}
P.~Ram and A.~G. Gray.
\newblock Maximum inner-product search using cone trees.
\newblock In {\em KDD}, pages 931--939, 2012.

\bibitem{Proc:Shrivastava_NIPS13}
A.~Shrivastava and P.~Li.
\newblock Beyond pairwise: Provably fast algorithms for approximate k-way
  similarity search.
\newblock In {\em NIPS}, Lake Tahoe, NV, 2013.

\bibitem{Proc:OneHashLSH_ICML14}
A.~Shrivastava and P.~Li.
\newblock Densifying one permutation hashing via rotation for fast near
  neighbor search.
\newblock In {\em ICML}, Beijing, China, 2014.

\bibitem{Proc:Weber_VLDB98}
R.~Weber, H.-J. Schek, and S.~Blott.
\newblock A quantitative analysis and performance study for similarity-search
  methods in high-dimensional spaces.
\newblock In {\em VLDB}, pages 194--205, 1998.

\end{thebibliography}
\end{document}